\documentclass[10pt,journal,compsoc]{IEEEtran}
\ifCLASSOPTIONcompsoc
  \usepackage[nocompress]{cite}
\else
  \usepackage{cite}
\fi
\ifCLASSINFOpdf
\usepackage[pdftex]{graphicx}
\else
\usepackage[dvips]{graphicx}
\fi
\usepackage{amsmath}
\usepackage{algorithmic}
\usepackage{amsmath}
\usepackage{amssymb}
\usepackage{hhline}
\usepackage{comment}
\usepackage{subfigure}
\usepackage{adjustbox}
\usepackage{multirow}
\usepackage{xcolor}
\usepackage{pbox}

\usepackage{array}
\newcolumntype{L}[1]{>{\raggedright\let\newline\\\arraybackslash\hspace{0pt}}m{#1}}
\newcolumntype{C}[1]{>{\centering\let\newline\\\arraybackslash\hspace{0pt}}m{#1}}
\newcolumntype{R}[1]{>{\raggedleft\let\newline\\\arraybackslash\hspace{0pt}}m{#1}}

\usepackage{url}

\newcommand{\ie}{i.e.}
\newcommand{\eg}{e.g.}

\begin{document}
%
\title{Open Set Domain Adaptation for Image and Action Recognition}
%
%
%
%

\author{Pau~Panareda~Busto,
        Ahsan~Iqbal,
        and~Juergen~Gall,~\IEEEmembership{Member,~IEEE}
\IEEEcompsocitemizethanks{\IEEEcompsocthanksitem P. Panareda Busto, M. Iqbal and J. Gall are with the Computer Vision Group, University of Bonn, Germany.\protect\\
E-mails: s6papana@uni-bonn.de, \{iqbalm,gall\}@iai.uni-bonn.de}
\thanks{Manuscript received January 29, 2018.}} 

%
%

\markboth{IEEE TRANSACTIONS ON PATTERN ANALYSIS AND MACHINE INTELLIGENCE,~VOL.~X,~NO.~X,~JANUARY~XXXX}%
{Shell \MakeLowercase{\textit{et al.}}: Bare Demo of IEEEtran.cls for Computer Society Journals}
%



\IEEEtitleabstractindextext{%
\begin{abstract}
Since annotating and curating large datasets is very expensive, there is a need to transfer the knowledge from existing annotated datasets to unlabelled data. 
Data that is relevant for a specific application, however, usually differs from publicly available datasets since it is sampled from a different domain. 
While domain adaptation methods compensate for such a domain shift, they assume that all categories in the target domain are known and match the categories in the source domain. 
Since this assumption is violated under real-world conditions, we propose an approach for open set domain adaptation where the target domain contains instances of categories that are not present in the source domain. The proposed approach achieves state-of-the-art results on various datasets for image classification and action recognition. Since the approach can be used for open set and closed set domain adaptation, as well as unsupervised and semi-supervised domain adaptation, it is a versatile tool for many applications.
\end{abstract}

}

\maketitle

\IEEEdisplaynontitleabstractindextext

\IEEEpeerreviewmaketitle

\IEEEraisesectionheading{\section{Introduction}\label{sec:introduction}}


\IEEEPARstart{I}{n} the last years, impressive results have been achieved on large-scale datasets for image classification or action recognition. Acquiring such large annotated datasets, however, is very expensive and there is a need to transfer the knowledge from existing annotated datasets to unlabelled data that is relevant for a specific application. If the labelled and unlabelled data have different characteristics, they have been sampled from two different domains. In particular, datasets that have been collected from the Internet, \eg, from platforms for sharing videos or images, differ greatly from data that needs to be processed for an application. To address the domain shift between the labelled dataset, which is the source domain, and the unlabelled data from the target domain, various unsupervised domain adaptation approaches have been proposed. If the data from the target source is partially labelled, the problem is termed semi-supervised domain adaptation. In this work, we address unsupervised and semi-supervised domain adaptation in the context of image and action recognition.         


\begin{figure}[t]
	\centering
	\subfigure[Closed set domain adaptation]{
		\includegraphics[width=1.0\linewidth]{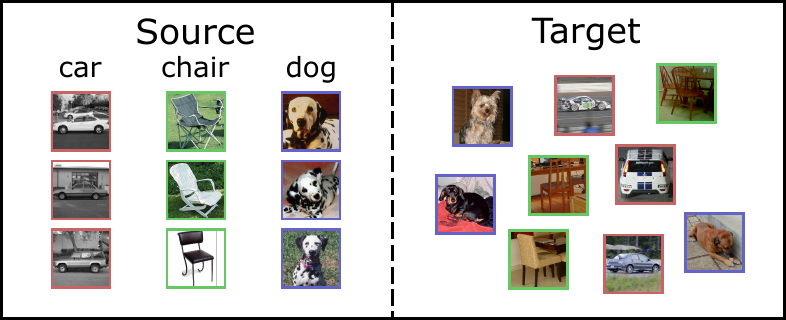}
		\label{fig:closed}
	}
	\subfigure[Open set domain adaptation]{
		\includegraphics[width=1.0\linewidth]{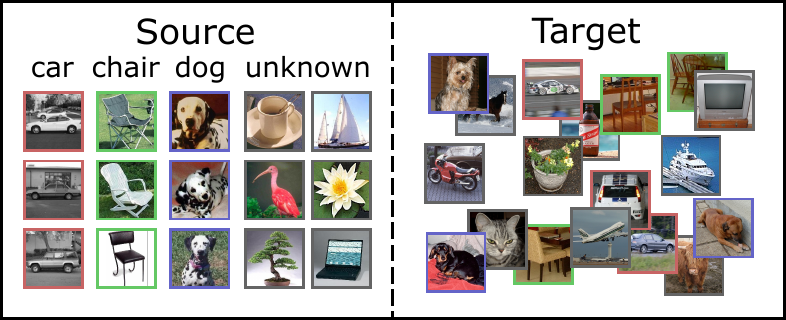}
		\label{fig:open}
	}
	\caption{(a) Standard domain adaptation benchmarks assume that source and target domains contain images or videos only of the same set of categories. This is denoted as \emph{closed set domain adaptation} since it does not include samples of unknown categories or categories which are not present in the other domain. (b) We propose \emph{open set domain adaptation}. In this setting, both source and target domain contain images or videos that do not belong to the categories of interest. Furthermore, the target domain contains images or videos that are not related to any image or video in the source domain and vice versa.         
	}
	\label{fig:teaser}
\end{figure}


Although the methods for domain adaptation have been advanced tremendously in the last years~\cite{DA_Saenko10,DA_Gopalan11,DA_Gong12,CNN-DA_Chopra13,DA_Hoffman14,CNN-DA_Ganin15,CNN-DA_Hsu15,CNN-DA_Ghifary16,CNN-DA_Tzeng17,CNN-DA_Motiian17}, the evaluation protocols were restricted to a scenario where all categories in the target domain are known and match the categories in the source domain. Fig.~\ref{fig:closed} illustrates such a \emph{closed set domain adaptation} setting. The assumption that all images or videos that are in the target domain belong to categories in the source domain, however, is violated in most cases. In particular if the number of potential categories is very large as it is the case for object or action categories, the target domain contains images or videos of categories that are not present in the source domain since they are not of interest for a specific application. We therefore propose a more realistic evaluation setting for unsupervised or semi-supervised domain adaptation, namely \emph{open set domain adaptation}, which builds on the concept of open sets~\cite{DATA_Scheirer13,DATA_Scheirer14,CNN_Bendale16}. As illustrated in Fig.~\ref{fig:teaser}, the source and target domains are not anymore restricted in the open set case to share the same categories as in the closed set case, but both domains contain images or videos from categories that are not present in the other domain.  

\begin{figure*}[t]
	\centering
	\subfigure[]{
		\includegraphics[width=0.185\linewidth]{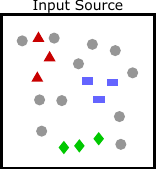}
		\includegraphics[width=0.185\linewidth]{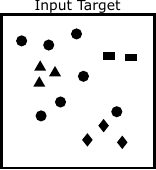}
		\label{fig:pipeline_input}
	}
	\subfigure[]{
		\includegraphics[width=0.185\linewidth]{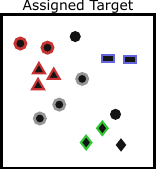}
		\label{fig:pipeline_assign}
	}
	\subfigure[]{
		\includegraphics[width=0.185\linewidth]{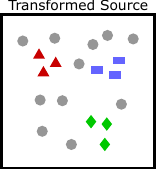}
		\label{fig:pipeline_transform}
	}
	\subfigure[\label{fig:pipeline_d}]{
		\includegraphics[width=0.187\linewidth]{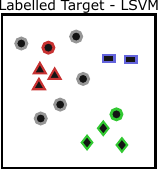}
		\label{fig:pipeline_lsvm}
	}
	\caption{Overview of the proposed approach for unsupervised \emph{open set domain adaptation}. (a) The source domain contains some labelled images, indicated by the colours red, blue and green, and some images belonging to unknown classes (grey). For the target domain, we do not have any labels but the shapes indicate if they belong to one of the three categories or an unknown category (circle). (b) In the first step, we assign class labels to some target samples, leaving outliers unlabelled. (c) By minimising the distance between the samples of the source and the target domain that are labelled by the same category, we learn a mapping from the source to the target domain. The image shows the samples in the source domain after the transformation. This process iterates between (b) and (c) until it converges to a local minimum. (d) In order to label all samples in the target domain either by one of the three classes (red, green, blue) or as unknown (grey), we learn a classifier on the source samples that have been mapped to the target domain (c) and apply it to the samples of the target domain (a). In this image, two samples with unknown classes are wrongly classified as red or green.}
	\label{fig:pipeline}
\end{figure*}

To address the problem of open set domain adaptation, we propose a generic approach that learns a linear mapping that maps the feature space of the source domain to the feature space of the target domain. It assigns a subset of images or videos of the target domain to the categories of the source domain and transforms the feature space of the source domain gradually towards the feature space of the target domain. By using a subset instead of the entire set, the approach handles images or videos in the target domain that are not related to any sample in the source domain. The approach can be applied to any feature space, which includes features extracted from images as well as features extracted from videos. The approach works in particular very well for features spaces that are extracted by convolutional networks and outperforms most end-to-end learning approaches for domain adaptation. The good performance of the approach coincides with the observation that deep convolutional networks tend to linearise manifolds of image domains~\cite{BengioMDR13,UpchurchGPPSBW17}. In this case, a linear mapping is sufficient to map the feature space of the source domain to the feature space of the target domain. In particular, the flexibility of the approach, which can be used for images and videos, for open set and closed set domain adaptation, as well as unsupervised and semi-supervised domain adaptation, makes the approach a versatile tool for applications. An overview of the approach for unsupervised open set domain adaptation is given in Fig.~\ref{fig:pipeline}. 

A preliminary version of this work was presented in~\cite{Busto17}. In this work, we introduce open set domain adaptation for action recognition and provide a thorough experimental evaluation, which includes open set domain adaptation from synthetic data to real data and an evaluation of the proposed approach for standard closed set protocols. In total, we evaluate the approach on 26 \emph{open set} and 34 \emph{closed set} combinations of source and target domains including the \emph{Office} dataset~\cite{DA_Saenko10}, its extension with the \emph{Caltech} dataset~\cite{DA_Gong12}, the \emph{Cross-Dataset Analysis}~\cite{DATA_Tommasi14}, the \emph{Sentiment dataset}~\cite{DA_Blitzer07}, synthetic data~\cite{DATA_Peng17}, and two action recognition datasets, namely the \emph{Kinetics Human Action Video Dataset}~\cite{Data_Kay17} and the \emph{UCF101 Action Recognition Dataset}~\cite{Data_Soomro12}.
Our approach achieves state-of-the-art results in all settings both for unsupervised and semi-supervised open set domain adaptation and obtains competitive results compared state-of-the-art deep leaning approaches for closed set domain adaptation. 

\section{Related Work}
\label{sec:related}

\subsection{Domain Adaptation}

The interest in studying domain adaptation techniques for computer vision problems increased with the release of a benchmark by Saenko et al.~\cite{DA_Saenko10} for domain adaptation in the context of object classification.
The first relevant works on unsupervised domain adaptation for object categorisation were presented by Golapan et al.~\cite{DA_Gopalan11} and Gong et al.~\cite{DA_Gong12}, who proposed an alignment in a common subspace of source and target samples using the properties of Grassmanian manifolds.
Jointly transforming source and target domains into a common low dimensional space was also done together with a conjugate gradient minimisation of a transformation matrix with orthogonality constraints~\cite{DA_Baktashmotlagh13} and with dictionary learning to find subspace interpolations~\cite{DA_Ni13,DA_Shekhar13,DA_Xu15}.
Sun et al.~\cite{DA_Sun14,CNN-DA_Sun15} presented a very efficient solution based on second-order statistics to align a source domain with a target domain.
Herath et al.~\cite{DA_Herath17} also match second-order statistics with a joint estimation of latent spaces. To obtain an estimate of the target distribution in the latent space, Gholami et al.~\cite{CNN-DA_Gholami17} introduce a Bayesian approximation to jointly learn a softmax classifier across-domains.
Similarly, Csurka et al.~\cite{DA_Csurka16} jointly denoise source and target samples to reconstruct data without partial random corruption.
Zhang et al.~\cite{DA_Zhang17} also align distributions, but they include geometrical differences in a joint optimisation.
Sharing certain similarities with associations between domains, Gong et al.~\cite{DA_Gong13} minimise the Maximum Mean Discrepancy (MMD)~\cite{MATH_Gretton06} of two datasets. They assign instances to latent domains and solve it by a relaxed binary optimisation.
Hsu et al.~\cite{CNN-DA_Hsu15} use a similar idea 
allowing instances to be linked to all other samples.

Semi-supervised domain adaptation approaches take advantage of knowing the class labels of a few target samples.
Aytar et al.~\cite{DA_Aytar11} proposed a transfer learning formulation to regularise the training of target classifiers.
Exploiting pairwise constraints across domains, Saenko et al.~\cite{DA_Saenko10} and Kulis et al.~\cite{DA_Kulis11} learn a transformation to minimise the effect of the domain shift while also training target classifiers.
Following the same idea, Hoffman et al.~\cite{DA_Hoffman13} considered an iterative process to alternatively minimise the classification weights and the transformation matrix. In a different context, \cite{DA_Busto15} proposed a weakly supervised approach to refine coarse viewpoint annotations of real images by synthetic images. In contrast to semi-supervised approaches, the task of viewpoint refinement assumes that all images in the target domain are labelled but not with the desired granularity.    

The idea of selecting the most relevant information of each domain has been studied in early domain adaptation methods in the context of natural language processing~\cite{DA_Blitzer06}. Pivot features that behave the same way for discriminative learning in both domains were selected to model their correlations.
Gong et al.~\cite{DA_Gong13ICML} presented an algorithm that selects a subset of source samples that are distributed most similarly to the target domain.
Another technique that deals with instance selection has been proposed by Sangineto et al.~\cite{DA_Sangineto14}.
They train weak classifiers on random partitions of the target domain and evaluate them in the source domain. The best performing classifiers are then selected.     
Other works have also exploited greedy algorithms that iteratively add target samples to the training process, while the least relevant source samples are removed~\cite{DA_Bruzzone10,CNN-DA_Tommasi13}. 

During the last years, a large number of domain adaptation methods have been based on deep convolutional neural networks (CNN)~\cite{CNN_Krizhevsky12}, which learn more discriminative feature representations than hand-crafted features and substantially reduce the domain bias between datasets in object recognition tasks~\cite{CNN_Donahue2014}.
Non-adapted classifiers trained with features extracted from CNN layers outperform domain adaptation methods with shallow feature descriptors~\cite{CNN_Donahue2014,CNN-DA_Sun15}. 
Many of these deep domain adaptation architectures are inspired by the traditional methods and seek to minimise the MMD distance as a regulariser to learn features for source and target samples jointly~\cite{CNN-DA_Ghifary14,CNN-DA_Tzeng14,CNN-DA_Long15,CNN-DA_Long16,CNN-DA_Yan17}.
Recently, Carlucci et al.~\cite{CNN-DA_Carlucci17} extend this type of networks and use intermediate layers for the alignment of distributions before batch normalisation. They learn a parameter that steers the contribution of each domain at a given layer.
Ganin et al.~\cite{CNN-DA_Ganin15} added a domain classifier network after the CNN to maximize the discriminatory loss of both domains while jointly minimising the classification loss using source data.
More recently, Tzeng et al.~\cite{CNN-DA_Tzeng17} propose a generalized framework for adversarial adaptation. 
In the semi-supervised setting, Mottian et al.~\cite{CNN-DA_Motiian17} present a deep domain adaptation method that exploits the domain loss minimisation while maximizing the distances between labelled samples from different domains and classes.
Other forms of data representation, such as hash codes~\cite{CNN-DA_Venkateswara17} and scatter tensors~\cite{CNN-DA_Koniusz17,DA_Lu17}, have also been combined with deep domain adaptation architectures to further reduce the domain bias.

\subsection{Open Set Recognition}
The inclusion of \emph{open sets} in recognition tasks appeared in the field of face recognition, where evaluation datasets contain unseen face instances as impostors that have to be rejected~\cite{DATA_Phillips00,DATA_Li05}. Such open set protocols are nowadays widely used for evaluating face recognition approaches~\cite{CNN_Sun15}. 

The generalisation towards an open set scenario for multi-object classification was introduced by Schreier et al.~\cite{DATA_Scheirer13}, who addressed the more realistic case of a finite set of known objects mixed with many unknown ones.
Based on this principle, \cite{OBJ_Jain14} and \cite{DATA_Scheirer14} propose multi-class classifiers that detect unknown instances by learning SVMs that assign probabilistic decision scores instead of class labels.
More recently, Bendale and Boult~\cite{CNN_Bendale16} adapt traditional neural networks for open set recognition tasks by introducing a new layer that estimates the probability of an object to be labelled as unseen class.

Closely related are also the works \cite{MATH_Zhang06} and \cite{MATH_Bartlett08} that add a regulariser to detect uninformative data and penalise a misclassification during training.
Lately, Gavves et al.~\cite{DA_Gavves15} present an active learning technique, whose intially trained SVMs on a subset of known classes are used as priors to further train novel object classes from other target datasets.

\section{Open Set Domain Adaptation}
\label{sec:method}

We present an approach that iterates between solving the labelling problem of target samples, \ie, associating a subset of the target samples to the known categories of the source domain, and computing a mapping from the source to the target domain by minimising the distances of the assignments. The transformed source samples are then used in the next iteration to re-estimate the assignments and update the transformation. This iterative process is repeated until convergence and is illustrated in Fig.~\ref{fig:pipeline}. 

In Section~\ref{sec:uns}, we describe the unsupervised assignment of target samples to categories of the source domain.
The semi-supervised case is described in Section~\ref{sec:sup}. 
Section~\ref{sec:transform} finally describes how the mapping from the source domain to the target domain is estimated from the previous assignments. This part is the same for the unsupervised and semi-supervised setting. 

\subsection{Unsupervised Domain Adaptation}
\label{sec:uns}

We first address the problem of unsupervised domain adaptation, \ie, none of the target samples are annotated, in an open set protocol.
Given a set of classes $\mathcal{C}$ in the source domain, including $|\mathcal{C}-1|$ known classes and an additional unknown class that gathers all instances from other irrelevant categories, we aim to label the target samples $\mathcal{T} = \{T_1,\dots,T_{|\mathcal{T}|}\}$ by a class $c \in \mathcal{C}$. 
We define the cost of assigning a target sample $T_t$ to a class $c$ by $d_{ct} = \left\Vert S_c - T_t \right\Vert^2_2$ where $T_t \in \mathbb{R}^D$ is the feature representation of the target sample $t$ and $S_c \in \mathbb{R}^D$ is the mean of all samples in the source domain labelled by class $c$. To increase the robustness of the assignment, we do not enforce that all target samples are assigned to a class as shown in Fig.~\ref{fig:pipeline}(b). The cost of declaring a target sample as outlier is defined by a parameter $\lambda$, which is discussed in Section~\ref{sec:experiments:lambda}.

Having defined the individual assignment costs, we can formulate the entire assignment problem by:  

\begin{equation}
\label{eq:assignments}
\begin{aligned}
& \underset{x_{ct},o_t}{\operatorname{\text{minimise}}}
& & \sum_{t}\bigg(\sum_{c} d_{ct} x_{ct} + \lambda o_t\bigg) \\
& \text{subject to} 
& & \sum_{c} x_{ct} + o_t = 1 & \forall t~\text{,} \\
& & & \sum_{t} x_{ct} \geq 1 & \forall c~\text{,} \\
& & & x_{ct}, o_t \in \{0,1\} & \forall c,t~\text{.}
\end{aligned}
\end{equation}
By minimising the constrained objective function, we obtain the binary variables $x_{ct}$ and $o_t$ as solution of the assignment problem. 
The first type of constraints ensures that a target sample is either assigned to one class, \ie, $x_{ct}=1$, or declared as outlier, \ie, $o_t=1$. The second type of constraints ensures that at least one target sample is assigned to each class \mbox{$c \in \mathcal{C}$}. We use the constraint integer program package SCIP~\cite{MATH_Achterberg09} to solve all proposed formulations. 

As it is shown in Fig.~\ref{fig:pipeline}(b), we label the targets also by the unknown class. Note that the unknown class combines all objects that are not of interest. Even if the unknowns in the source and target domain belong to different semantic classes, a target sample might be closer to the mean of all negatives than to any other positive class. In this case, we can confidentially label a target sample as unknown.                  
In our experiments, we show that it makes not much difference if the unknown class is included in the unsupervised setting since the outlier handling discards target samples that are not close to the mean of negatives. 

\subsection{Semi-supervised Domain Adaptation}
\label{sec:sup}

The unsupervised assignment problem naturally extends to a semi-supervised setting when a few target samples are annotated. In this case, we only have to extend the formulation \eqref{eq:assignments} by additional constraints that enforce that the annotated target samples do not change the label, \ie,
\begin{equation}\label{eq:sup}
~~~~~~~~~~~~~~~~~~~~~~x_{\hat{c}_tt} = 1 ~~~~~~~~~~~~~~~~~~~~~~~~\forall (t,\hat{c}_t) \in \mathcal{L} \text{,}
\end{equation}
where $\mathcal{L}$ denotes the set of labelled target samples and $\hat{c}_t$ the class label provided for target sample $t$.
In order to exploit the labelled target samples better, one can use the neighbourhood structure in the source and target domain. 
While the constraints remain the same, the objective function \eqref{eq:assignments} can be changed to
\begin{equation}
\label{eq:lc}
\begin{aligned}
& 
& & \sum_{t}\left(\sum_{c} x_{ct} \bigg( d_{ct} + \sum_{t' \in N_t} \sum_{c'} d_{cc'} x_{c't'} \bigg) + \lambda o_t \right) \text{,} \\
\end{aligned}
\end{equation}
where $ d_{cc'} = \left\Vert S_c - S_{c'} \right\Vert^2_2$. While in \eqref{eq:assignments} the cost of labelling a target sample $t$ by the class $c$ is given only by $d_{ct}$, a second term is added in \eqref{eq:lc}. It is computed over all neighbours $N_t$ of $t$ and adds the distance between the classes in the source domain as additional cost if a neighbour is assigned to another class than the target sample $t$. 

The objective function \eqref{eq:lc}, however, becomes quadratic and therefore NP-hard to solve.
Thus, we transform the \emph{quadratic assignment problem} into a mixed 0-1 linear program using the Kaufman and Broeckx linearisation~\cite{MATH_Kaufman78}. 
By substituting
\begin{equation}\label{eq:L}
w_{ct} =  x_{ct} \left( \sum_{t' \in N_t} \sum_{c'}  d_{cc'} x_{c't'} \right)  \text{,}
\end{equation}
we derive to the linearised problem
\begin{equation}
\label{eq:linearized}
\begin{aligned}
& \underset{x_{ct},w_{ct},o_t}{\operatorname{\text{minimise}}}
& & \sum_{t}\left(\sum_{c} d_{ct} x_{ct} + \sum_{c} w_{ct} + \lambda o_t\right)\\
& \text{subject to}
& & \sum_{c} x_{ct} + o_t = 1 & \forall t~\text{,} \\
& & & \sum_{t} x_{ct} \geq 1 & \forall c~\text{,} \\
& & & a_{ct}x_{ct}+\sum_{t' \in N_t} \sum_{c'} d_{cc'} x_{c't'} - w_{ct} \le a_{ct} & \forall s,t~\text{,} \\
& & & x_{ct}, o_t \in \{0,1\} & \forall c,t~\text{,} \\
& & & w_{ct} \ge 0 & \forall c,t~\text{,} 
\end{aligned}
\end{equation}
where $a_{ct} = \sum_{t' \in N_t} \sum_{c'} d_{cc'}$. 

\subsection{Mapping}
\label{sec:transform}

As illustrated in Fig.~\ref{fig:pipeline}, we iterate between solving the assignment problem, as described in Section \ref{sec:uns} or \ref{sec:sup}, and estimating the mapping from the source domain to the target domain.     
We consider a linear transformation, which is represented by a matrix $W \in \mathbb{R}^{D \times D}$. We estimate $W$ by minimising the following loss function:
\begin{equation}\label{eq:objMat}
f(W) = \dfrac{1}{2}\sum_t\sum_c x_{ct} \Vert W S_c - T_t \Vert^2_2~\text{,}
\end{equation}
which can be written in matrix form:
\begin{equation}\label{eq:objMat2}
f(W) = \dfrac{1}{2}|| W P_S - P_T ||^2_F~\text{.}
\end{equation}
The matrices $P_S$ and $P_T \in \mathbb{R}^{D\times L}$ with $L=\sum_t\sum_c x_{ct}$ represent all assignments, where the columns denote the actual associations.
The quadratic nature of the convex objective function may be seen as a linear least squares problem, which can be easily solved by any available QP solver. State-of-the-art features based on convolutional neural networks, however, are high dimensional and the number of target instances is usually very large. We use therefore non-linear optimisation~\cite{MATH_Svanberg02,MATH_Johnson10} to optimise $f(W)$. The derivatives of \eqref{eq:objMat} are given by
\begin{equation}\label{eq:diff}
\dfrac{\partial f(W)}{\partial W} = W(P_SP_S^T) - P_TP_S^T~\text{.}
\end{equation}
If $L < D$, \ie, the number of samples, which have been assigned to a known class, is smaller than the dimensionality of the features, the optimisation also deals with an underdetermined linear least squares formulation. In this case, the solver converges to the matrix $W$ with the smallest norm, which is still a valid solution.

After the transformation $W$ is estimated, we map the source samples to the target domain.
We therefore iterate the process of solving the assignment problem and estimating the mapping from the source domain to the target domain until it converges. 
After the approach has converged, we train linear SVMs in a one-vs-one setting on the transformed source samples. For the semi-supervised setting, we also include the annotated target samples $\mathcal{L}$ \eqref{eq:sup} to the training set. The linear SVMs are then used to obtain the final labelling of the target samples as illustrated in Fig.~\ref{fig:pipeline_d}. 

\section{Experiments}
\label{sec:experiments}

We evaluate our method in the context of domain adaptation for image classification and action recognition. 
In this setting, the images or videos of the source domain are annotated by class labels and the goal is to classify the images or videos in the target domain. We report the accuracies for both unsupervised and semi-supervised scenarios, where target samples are unlabelled or partially labelled, respectively. 
For consistency, we use \emph{libsvm}~\cite{MATH_Chang01} 
since it has also been used in other works, \eg, \cite{DA_Fernando13} and \cite{CNN-DA_Sun15}.
We set the misclassification parameter $C = 0.001$ in all experiments, which allows for a soft margin optimisation that works best in such classification tasks~\cite{DA_Fernando13,CNN-DA_Sun15}. The source code and the described open set protocols are available at \url{https://github.com/Heliot7/open-set-da}. 

\subsection{Parameter configuration}
\label{sec:experiments:lambda}

Our algorithm contains a few parameters that need to be defined.
For the outlier rejection, we use
\begin{equation}\label{eq:lambda}
\lambda = \rho\big(\max_{t,c}d_{ct} + \min_{t,c}d_{ct}\big), 
\end{equation}
\ie, $\lambda$ is adapted automatically based on the distances $d_{ct}$ and $\rho$, which 
is set to $0.5$ unless otherwise specified.
While higher values of $\lambda$ closer to the largest distance barely discard any outlier, lower values almost reject all assignments.
We iterate the approach until the maximum number of 10 iterations is reached or if the distance  
\begin{equation}\label{eq:diststop}
\sqrt{\sum_{t}\sum_{c}x_{ct}\left\Vert W_k S_{c} - T_{t}\right\Vert_2^2}~~\text{}
\end{equation}
is below $\epsilon = 0.01$, where $W_{k}$ denotes the estimated transformation 
at iteration $k$. 
In practice, the process converges after 3-5 iterations. 

\subsection{Open set domain adaptation}

\subsubsection{Office dataset}
\label{exp:office}

We evaluate and compare our approach on the \emph{Office} dataset~\cite{DA_Saenko10}, which is the standard benchmark for domain adaptation with CNN features. It provides three different domains, namely \emph{Amazon (A)}, \emph{DSLR (D)} and \emph{Webcam (W)}. While the \emph{Amazon} dataset contains centred objects on white background, the other two comprise pictures taken in an office environment but with different quality levels.
In total, there are 31 common classes for 6 source-target combinations. This means that there are 4 combinations with a considerable domain shift (A $\rightarrow$ D, A $\rightarrow$ W, D $\rightarrow$ A, W $\rightarrow$ A) and 2 with a minor domain shift (D $\rightarrow$ W, W $\rightarrow$ D). 
{Following the standard protocol and for a fair comparison with the other methods, we extract feature vectors from the fully connected layer-7 (fc7) of the AlexNet model~\cite{CNN_Krizhevsky12}}. 
	
We introduce an open set protocol for this dataset by taking the 10 classes that are also common in the \emph{Caltech} dataset~\cite{DA_Gong12} as shared classes. In alphabetical order, the classes 11-20 are used as unknowns in the source domain and 21-31 as unknowns in the target domain, \ie, the unknown classes in the source and target domain are not shared. For evaluation, each sample in the target domain needs to be correctly classified either by one of the 10 shared classes or as unknown. In order to compare with a closed setting (CS), we report the accuracy when source and target domain contain only samples of the 10 shared classes. Since OS is evaluated on all target samples, we also report the numbers when the accuracy is only measured on the same target samples as CS, \ie, only for the shared 10 classes. The latter protocol is denoted by OS$^*$(10) and provides a direct comparison to CS(10). 

\begin{table}[t]
	\scriptsize
	\setlength{\tabcolsep}{0.175em}
	\begin{tabular}{|l|c|c|c|c|c|c|}
		\cline{2-7}
		\multicolumn{1}{c|}{} &
		\multicolumn{3}{c|}{A$\rightarrow$D} & \multicolumn{3}{c|}{A$\rightarrow$W} \\ \cline{2-7}
		\multicolumn{1}{c|}{} &
		CS (10) & OS$^*$ (10) & OS (10) & CS (10) & OS$^*$ (10) & OS (10) \\ \hhline{|=|=|=|=|=|=|=|}
		LSVM & 87.1 & 70.7 & 72.6 & 77.5 & 53.9 & 57.5 \\ \hhline{|=|=|=|=|=|=|=|}
		DAN~\cite{CNN-DA_Long15} & 88.1 & 76.5 & 77.6 & \textbf{90.5} & 70.2 & 72.5 \\ \hline
		RTN~\cite{CNN-DA_Long16} & \textbf{93.0} & 74.7 & 76.6 & 87.0 & 70.8 & 73.0 \\ \hline
		BP~\cite{CNN-DA_Ganin15} & 91.9 & 77.3 & 78.3 & 89.2 & 73.8 & 75.9 \\ \hhline{|=|=|=|=|=|=|=|}
		ATI & 92.4 & 78.2 & 78.8 & 85.1 & \textbf{77.7} & \textbf{78.4} \\ \hline
		ATI-$\lambda$ & \textbf{93.0} & \textbf{79.2} & \textbf{79.8} & 84.0 & 76.5 & 77.6 \\ \hline
		ATI-$\lambda$-N1 & 91.9 & 78.3 & 78.9 & 84.6 & 74.2 & 75.6 \\ \hline
	\end{tabular}
	\begin{tabular}{|l|c|c|c|c|c|c|}
		\cline{2-7}
		\multicolumn{1}{c|}{} &
		\multicolumn{3}{c|}{D$\rightarrow$A} & \multicolumn{3}{c|}{D$\rightarrow$W} \\ \cline{2-7}
		\multicolumn{1}{c|}{} &
		CS (10) & OS$^*$ (10) & OS (10) & CS (10) & OS$^*$ (10) & OS (10) \\ \hhline{|=|=|=|=|=|=|=|}
		LSVM & 79.4 & 40.0 & 45.1 & 97.9 & 87.5 & 88.5 \\ \hhline{|=|=|=|=|=|=|=|}
		DAN~\cite{CNN-DA_Long15} & 83.4 & 53.5 & 57.0 & 96.1 & 87.5 & 88.4 \\ \hline
		RTN~\cite{CNN-DA_Long16} & 82.8 & 53.8 & 57.2 & 97.9 & 88.1 & 89.0 \\ \hline
		BP~\cite{CNN-DA_Ganin15} & 84.3 & 54.1 & 57.6 & 97.5 & 88.9 & 89.8 \\ \hhline{|=|=|=|=|=|=|=|}
		ATI & 93.4 & \textbf{70.0} & 71.1 & \textbf{98.5} & 92.2 & 92.6 \\ \hline
		ATI-$\lambda$ & \textbf{93.8} & \textbf{70.0} & \textbf{71.3} & \textbf{98.5} & 93.2 & 93.5 \\ \hline
		ATI-$\lambda$-N1 & 93.3 & 65.6 & 67.8 & 97.9 & \textbf{94.0} & \textbf{94.4} \\ \hline
	\end{tabular}
	\begin{tabular}{|l|c|c|c|c|c|c|c|c|c|}
		\cline{2-10}
		\multicolumn{1}{c|}{} &
		\multicolumn{3}{c|}{W$\rightarrow$A} & \multicolumn{3}{c|}{W$\rightarrow$D} & \multicolumn{3}{c|}{AVG.} \\ \cline{2-10}
		\multicolumn{1}{c|}{} & 
		CS (10) & OS$^*$ (10) & OS (10) & CS (10) & OS$^*$ (10) & OS (10) & CS & OS$^*$ & OS \\ \hhline{|=|=|=|=|=|=|=|=|=|=|}
		LSVM & 80.0 & 44.9 & 49.2 & \textbf{100} & 96.5 & 96.6 & 87.0 & 65.6 & 68.3 \\ \hhline{|=|=|=|=|=|=|=|=|=|=|}
		DAN~\cite{CNN-DA_Long15} & 84.9 & 58.5 & 60.8 & \textbf{100} & 97.5 & 98.3 & 90.5 & 74.0 & 75.8 \\ \hline
		RTN~\cite{CNN-DA_Long16} & 85.1 & 60.2 & 62.4 & \textbf{100} & 98.3 & \textbf{98.8} & 91.0 & 74.3 & 76.2 \\ \hline
		BP~\cite{CNN-DA_Ganin15} & 86.2 & 61.8 & 64.0 & \textbf{100} & 98.0 & 98.7 & 91.6 & 75.7 & 77.4 \\ \hhline{|=|=|=|=|=|=|=|=|=|=|}
		ATI & 93.4 & 76.4 & 76.6 & \textbf{100} & 99.1 & 98.3 & \textbf{93.8} & 82.1 & 82.6 \\ \hline
		ATI-$\lambda$ & \textbf{93.7} & \textbf{76.5} & \textbf{76.7} & \textbf{100} & 99.2 & 98.3 & 93.7 & \textbf{82.4} & \textbf{82.9} \\ \hline
		ATI-$\lambda$-N1 & 93.4 & 71.6 & 72.4 & \textbf{100} & \textbf{99.6} & \textbf{98.8} & 93.5 & 80.6 & 81.3 \\ \hline
	\end{tabular}
	\vspace{1.25mm}
	\caption{Open set domain adaptation on the unsupervised Office dataset with 10 shared classes (OS) using all samples per class~\cite{DA_Gong13}. For comparison, results for closed set domain adaptation (CS) and modified open set (OS$^*$) are reported.   
	}
	\vspace{-2.5mm}
	\label{table:office_uns_ft}
\end{table}


\begin{table*}[h]
	\scriptsize
	\begin{center}
		\setlength{\tabcolsep}{.375em}
		\begin{tabular}{|l|c|c|c|c|c|c|c|c|c|c|c|c|}
			\cline{2-13}
			\multicolumn{1}{c|}{} & \multicolumn{2}{c|}{A$\rightarrow$D} & \multicolumn{2}{c|}{A$\rightarrow$W} & \multicolumn{2}{c|}{D$\rightarrow$A} &\multicolumn{2}{c|}{D$\rightarrow$W} & \multicolumn{2}{c|}{W$\rightarrow$A} & \multicolumn{2}{c|}{W$\rightarrow$D} \\ \cline{2-13}
			\multicolumn{1}{c|}{} & assign-$\lambda$ & LSVM & assign-$\lambda$ & LSVM & assign-$\lambda$ & LSVM & assign-$\lambda$ & LSVM & assign-$\lambda$ & LSVM & assign-$\lambda$ & LSVM \\ \hline
			initial     &      & 72.6 &      & 57.5 &      & 45.1 &      & 88.5 &      & 49.2 &      & 96.6 \\ \hline
			iteration 1 & 78.4 & 76.8 & 74.5 & 69.8 & 73.6 & 68.1 & 90.4 & 90.3 & 71.9 & 70.0 & 89.6 & 97.8 \\ \hline
			iteration 2 & 77.7 & 79.1 & 80.1 & 77.6 & 80.4 & 71.3 & 91.5 & 93.5 & 77.2 & 75.9 & 84.7 & 98.3 \\ \hline
			iteration 3 & 75.3 & 79.8 &      &      &      &      &      &      & 77.8 & 76.7 &      &  \\ \hline
		\end{tabular}
	\end{center}
	\caption{Evolution of the percentage of correct assignments (assign-$\lambda$) when taking into account the selected target samples and the average class accuracy of all target samples using linear SVMs (LSVM). The approach converges after 2 or 3 iterations.}
	\label{table:accuracy_assigns}
\end{table*}

\begin{table}[t]
	\scriptsize
	\setlength{\tabcolsep}{0.15em}
	\begin{tabular}{|l|c|c|c|c|c|c|}
		\cline{2-7}
		\multicolumn{1}{c|}{} &
		\multicolumn{3}{c|}{A$\rightarrow$D} & \multicolumn{3}{c|}{A$\rightarrow$W} \\ \cline{2-7}
		\multicolumn{1}{c|}{} &
		CS (10) & OS$^*$ (10) & OS (10) & CS (10) & OS$^*$ (10) & OS (10) \\ \hhline{|=|=|=|=|=|=|=|}
		LSVM & 84.4$\pm$5.9 & 63.7$\pm$6.7 & 66.6$\pm$5.9 & 76.5$\pm$2.9 & 48.2$\pm$4.8 & 52.5$\pm$4.2 \\ \hhline{|=|=|=|=|=|=|=|}
		TCA~\cite{DA_Pan09} & 85.9$\pm$6.3 & 75.5$\pm$6.6 & 75.7$\pm$5.9 & 80.4$\pm$6.9 & 67.0$\pm$5.9 & 67.9$\pm$5.5 \\ \hline
		gfk~\cite{DA_Gong12} & 84.8$\pm$5.1 & 68.6$\pm$6.7 & 70.4$\pm$6.0 & 76.7$\pm$3.1 & 54.1$\pm$4.8 & 57.4$\pm$4.2 \\ \hline
		SA~\cite{DA_Fernando13} & 84.0$\pm$3.4 & 71.5$\pm$5.9 & 72.6$\pm$5.3 & 76.6$\pm$2.8 & 57.4$\pm$4.2 & 60.1$\pm$3.7\\ \hline
		CORAL~\cite{CNN-DA_Sun15} & 85.8$\pm$7.2 & 79.9$\pm$5.7 & 79.6$\pm$5.0 & 81.9$\pm$2.8 & 68.1$\pm$3.6 & 69.3$\pm$3.1 \\ \hhline{|=|=|=|=|=|=|=|}
		ATI & \textbf{91.4$\pm$1.3} & 80.5$\pm$2.0 & 81.1$\pm$2.8 & \textbf{86.1$\pm$1.1} & 73.4$\pm$2.0 & \textbf{75.3$\pm$1.7} \\ \hline
		ATI-$\lambda$ & 91.1$\pm$2.1 & \textbf{81.1$\pm$0.4} & \textbf{82.2$\pm$2.0} & 85.5$\pm$2.1 & \textbf{73.7$\pm$2.6} & \textbf{75.3$\pm$1.4} \\ \hline
	\end{tabular}
	\begin{tabular}{|l|c|c|c|c|c|c|}
		\cline{2-7}
		\multicolumn{1}{c|}{} &
		\multicolumn{3}{c|}{D$\rightarrow$A} & \multicolumn{3}{c|}{D$\rightarrow$W} \\ \cline{2-7}
		\multicolumn{1}{c|}{} &
		CS (10) & OS$^*$ (10) & OS (10) & CS (10) & OS$^*$ (10) & OS (10) \\ \hhline{|=|=|=|=|=|=|=|}
		LSVM & 75.5$\pm$2.1 & 36.1$\pm$3.7 & 42.2$\pm$3.3 & 96.2$\pm$1.0 & 81.5$\pm$1.5 & 83.1$\pm$1.3 \\ \hhline{|=|=|=|=|=|=|=|}
		TCA~\cite{DA_Pan09} & 88.2$\pm$1.5 & 71.8$\pm$2.5 & 71.8$\pm$2.0 & 97.8$\pm$0.5 & 92.0$\pm$0.9 & 91.5$\pm$1.0 \\ \hline
		gfk~\cite{DA_Gong12} & 79.7$\pm$1.0 & 45.3$\pm$3.7 & 49.7$\pm$3.4 & 96.3$\pm$0.9 & 85.1$\pm$2.7 & 86.2$\pm$2.4 \\ \hline
		SA~\cite{DA_Fernando13} & 81.7$\pm$0.7 & 52.5$\pm$3.0 & 55.8$\pm$2.7 & 96.3$\pm$0.8 & 86.8$\pm$2.5 & 87.7$\pm$2.3 \\ \hline
		CORAL~\cite{CNN-DA_Sun15} & 89.6$\pm$1.0 & 66.6$\pm$2.8 & 68.2$\pm$2.5 & 97.2$\pm$0.7 & 91.1$\pm$1.7 & 91.4$\pm$1.5\\ \hhline{|=|=|=|=|=|=|=|}
		ATI & 93.5$\pm$0.3 & 69.8$\pm$1.4 & 70.8$\pm$2.1 & 97.3$\pm$0.5 & 89.6$\pm$2.1 & 90.3$\pm$1.8 \\ \hline
		ATI-$\lambda$ & \textbf{93.9$\pm$0.4} & \textbf{71.1$\pm$0.9} & \textbf{72.0$\pm$0.5} & \textbf{97.5$\pm$1.1} & \textbf{92.1$\pm$1.3} & \textbf{92.5$\pm$0.7} \\ \hline
	\end{tabular}
	\begin{tabular}{|l|c|c|c|c|c|c|c|c|c|}
		\cline{2-10}
		\multicolumn{1}{c|}{} &
		\multicolumn{3}{c|}{W$\rightarrow$A} & \multicolumn{3}{c|}{W$\rightarrow$D} & \multicolumn{3}{c|}{AVG.} \\ \cline{2-10}
		\multicolumn{1}{c|}{} & 
		CS (10) & OS$^*$ (10) & OS (10) & CS (10) & OS$^*$ (10) & OS (10) & CS & OS$^*$ & OS \\ \hhline{|=|=|=|=|=|=|=|=|=|=|}
		LSVM & 72.5$\pm$2.7 & 34.3$\pm$4.9 & 39.9$\pm$4.4 & 99.1$\pm$0.5 & 89.8$\pm$1.5 & 90.5$\pm$1.3 & 84.1 & 58.9 & 62.5 \\ \hhline{|=|=|=|=|=|=|=|=|=|=|}
		TCA & 85.5$\pm$3.3 & 68.1$\pm$5.1 & 68.6$\pm$4.6 & 98.8$\pm$0.9 & 94.1$\pm$2.9 & 93.6$\pm$2.6 & 89.5 & 78.1 & 78.2 \\ \hline
		gfk & 75.0$\pm$2.9 & 43.2$\pm$5.1 & 47.6$\pm$4.6 & 99.0$\pm$0.5 & 92.0$\pm$1.5 & 92.2$\pm$1.4 & 85.2 & 64.7 & 67.3 \\ \hline
		SA & 76.5$\pm$3.2 & 49.7$\pm$5.1 & 53.0$\pm$4.6 & 98.8$\pm$0.7 & 92.4$\pm$2.9 & 92.4$\pm$2.8 & 85.7 & 68.4 & 70.3 \\ \hline
		CORAL & 86.9$\pm$1.9 & 63.9$\pm$4.9 & 65.6$\pm$4.3 & \textbf{99.2$\pm$0.7} & 96.0$\pm$2.1 & 95.0$\pm$2.0 & 90.1 & 77.6 & 78.2 \\ \hhline{|=|=|=|=|=|=|=|=|=|=|}
		ATI & 92.2$\pm$1.1 & 75.1$\pm$1.7 & 76.0$\pm$2.0 & 98.9$\pm$1.3 & 95.5$\pm$2.3 & 95.4$\pm$2.1 & \textbf{93.2} & 80.7 & 81.5 \\ \hline
		ATI-$\lambda$ & \textbf{92.4$\pm$1.1} & \textbf{75.4$\pm$1.8} & \textbf{76.4$\pm$1.8} & 98.9$\pm$1.3 & \textbf{96.5$\pm$2.1} & \textbf{95.8$\pm$1.8} & \textbf{93.2} & \textbf{81.5} & \textbf{82.3} \\ \hline
	\end{tabular}
	\vspace{2.5mm}
	\caption{Open set domain adaptation on the unsupervised Office dataset with 10 shared classes (OS). We report the average and the standard deviation using a subset of samples per class in 5 random splits~\cite{DA_Saenko10}. For comparison, results for closed set domain adaptation (CS) and modified open set (OS$^*$) are reported. }
	\vspace{-2.5mm}
	\label{table:office_uns_it}
\end{table}

\vspace{1mm}\noindent{\bf Unsupervised domain adaptation.}
We firstly compare the accuracy of our method in the unsupervised set-up with state-of-the-art domain adaptation techniques embedded in the training of CNN models. 
DAN~\cite{CNN-DA_Long15} retrains the AlexNet model by freezing the first 3 convolutional layers, finetuning the last 2 and learning the weights from each fully connected layer by also minimising the discrepancy between both domains.
RTN~\cite{CNN-DA_Long16} extends DAN by adding a residual transfer module that bridges the source and target classifiers.
BP~\cite{CNN-DA_Ganin15} trains a CNN for domain adaptation by a gradient reversal layer and minimises the domain loss jointly with the classification loss. For training, we use all samples per class as proposed in~\cite{DA_Gong13}, which is the standard protocol for CNNs on this dataset. 
As proposed in~\cite{CNN-DA_Ganin15}, we use for all methods linear SVMs for classification instead of the soft-max layer for a fair comparison.  


To analyse the formulations that are discussed in Section~\ref{sec:method}, we compare several variants:    
ATI (\emph{Assign-and-Transform-Iteratively}) denotes our formulation in \eqref{eq:assignments} assigning a source class to all target samples, \ie, $\lambda = \infty$.
Then, ATI-$\lambda$ includes the outlier rejection and ATI-$\lambda$-$N_1$ is the unsupervised version of the locality constrained formulation corresponding to \eqref{eq:lc} with 1 nearest neighbour.
In addition, we denote LSVM as the linear SVMs trained on the source domain without any domain adaptation. 

\begin{table*}[t]
	\scriptsize
	\begin{center}
		\setlength{\tabcolsep}{.375em}
		\begin{tabular}{|l|c|c|c|c|c|c|c|c|c|c|c|c|c|c|}
			\cline{2-15}
			\multicolumn{1}{c|}{} &
			\multicolumn{2}{c|}{A$\rightarrow$D} & \multicolumn{2}{c|}{A$\rightarrow$W} & \multicolumn{2}{c|}{D$\rightarrow$A} & \multicolumn{2}{c|}{D$\rightarrow$W} & \multicolumn{2}{c|}{W$\rightarrow$A} & \multicolumn{2}{c|}{W$\rightarrow$D} & \multicolumn{2}{c|}{AVG.} \\ \cline{2-15}
			\multicolumn{1}{c|}{} & OS-SVM & LSVM & OS-SVM & LSVM & OS-SVM & LSVM & OS-SVM & LSVM & OS-SVM & LSVM & OS-SVM & LSVM & OS-SVM & LSVM \\ \hline
			
			No Adap. & 67.5 & 72.6 & 58.4 & 57.5 & 54.8 & 45.1 & 80.0 & 88.5 & 55.3 & 49.2 & 94.0 & 96.6 & 68.3 & 68.3 \\ \hline
			ATI-$\lambda$ & 72.0 & \textbf{79.8} & 65.3 & \textbf{77.6} & 66.4 & \textbf{71.3} & 82.2 & \textbf{93.5} & 71.6 & \textbf{76.7} & 92.7 & \textbf{98.3} & 75.0 & \textbf{82.9} \\ \hline 
		\end{tabular}
	\end{center}
	\caption{Comparison of a standard linear SVM (LSVM) with a specific open set SVM (OS-SVM)~\cite{DATA_Scheirer13} on the unsupervised Office dataset with 10 shared classes using all samples per class~\cite{DA_Gong13}. }
	\label{table:office_wsvm}
\end{table*}

\begin{table*}[t]
	\scriptsize
	\begin{center}
		\setlength{\tabcolsep}{.375em}
		\begin{tabular}{|l|c|c|c|c|c|c|c|}
			\cline{2-8}
			\multicolumn{1}{c|}{} &
			\multicolumn{1}{c|}{A$\rightarrow$D} & \multicolumn{1}{c|}{A$\rightarrow$W} & \multicolumn{1}{c|}{D$\rightarrow$A} & \multicolumn{1}{c|}{D$\rightarrow$W} & \multicolumn{1}{c|}{W$\rightarrow$A} & \multicolumn{1}{c|}{W$\rightarrow$D} & \multicolumn{1}{c|}{AVG.} \\ \cline{2-8}
			\multicolumn{1}{c|}{} & \multicolumn{7}{c|}{OS(10)} \\ \hline
			ATI-$\lambda$ ($\mathcal{C}$ w/o unknown) & 79.0 & 77.1 & 70.5 & 93.4 & 75.8 & 98.2 & 82.3 \\ \hline
			ATI-$\lambda$ ($\mathcal{C}$ with unknown) & \textbf{79.8} & \textbf{77.6} & \textbf{71.3} & \textbf{93.5} & \textbf{76.7} & \textbf{98.3} & \textbf{82.9} \\ \hline
		\end{tabular}
	\end{center}
	\caption{Impact of including the unknown class to the set of classes $\mathcal{C}$. The evaluation is performed on the unsupervised Office dataset with 10 shared classes using all samples per class~\cite{DA_Gong13}.} 
	\label{table:office_W}
\end{table*}

The results of these techniques using the described open set protocol are shown in Table~\ref{table:office_uns_ft}. Our approach ATI improves over the baseline without domain adaptation (LSVM) by +6.8\% for CS and +14.3\% for OS. The improvement is larger for the combinations that have larger domain shifts, \ie, the combinations that include the $Amazon$ dataset. We also observe that ATI outperforms all CNN-based domain adaptation methods for the closed (+2.2\%) and open setting (+5.2\%). It can also be observed that the accuracy for the open set is lower than for the closed set for all methods, but that our method handles the open set protocol best. While ATI-$\lambda$ does not obtain any considerable improvement compared to ATI in CS, the outlier rejection allows for an improvement in OS. The locality constrained formulation, ATI-$\lambda$-$N_1$, which we propose only for the semi-supervised setting, decreases the accuracy in the unsupervised setting.

{
The evolution of the percentage of correct assignments and the intermediate classification accuracies are shown in Table~\ref{table:accuracy_assigns}. The approach converges after two or three iterations. While the accuracy of the LSVMs that are trained on the transformed source samples increases with each iteration, the accuracy of the assignment can even decrease in some cases.         
}

Additionally, we report accuracies of popular domain adaptation methods that are not related to deep learning.
We report the results of methods that transform the data to a common low dimensional subspace, including Transfer Component Analysis (TCA)~\cite{DA_Pan09}, Geodesic Flow Kernel (GFK)~\cite{DA_Gong12} and Subspace alignment (SA)~\cite{DA_Fernando13}. In addition, we also include CORAL~\cite{CNN-DA_Sun15}, which whitens and recolours the source towards the target data.
Following the standard protocol of~\cite{DA_Saenko10}, we take 20 samples per object class when \emph{Amazon} is used as source domain, and 8 for \emph{DSLR} or \emph{Webcam}.
{As in the previous comparison with the CNN-based methods, we extract feature vectors from the last convolutional layer (fc7) from the AlexNet model~\cite{CNN_Krizhevsky12}}.
Each evaluation is executed 5 times with random samples from the source domain.
The average accuracy and standard deviation of the five runs are reported in Table~\ref{table:office_uns_it}. The results are similar to the protocol reported in Table~\ref{table:office_uns_ft}. Our approach ATI outperforms the other methods both for CS and OS and the additional outlier handling (ATI-$\lambda$) does not improve the accuracy for the closed set but for the open set.
 
\vspace{1mm}\noindent{\bf Impact of unknown class.}
The linear SVM that we employ in the open set protocol uses the unknown classes of the transformed source domain for the training. Since unknown object samples from the source domain are from different classes than the ones from the target domain, using an SVM that does not require any negative samples might be a better choice. 
Therefore, we compare the performance of a standard SVM classifier with a specific open set SVM (OS-SVM)~\cite{DATA_Scheirer14}, where only the 10 known classes are used for training. OS-SVM introduces an inclusion probability and labels target instances as unknown if this inclusion is not satisfied for any class.
Table~\ref{table:office_wsvm} compares the classification accuracies of both classifiers in the 6 domain shifts of the Office dataset. 
While the performance is comparable when no domain adaptation is applied, ATI-$\lambda$ obtains significantly better accuracies when the learning includes negative instances.

As discussed in Section~\ref{sec:uns}, the unknown class is also part of the labelling set $\mathcal{C}$ for the target samples. The labelled target samples are then used to estimate the mapping $W$~(\ref{eq:objMat}). To evaluate the impact of including the unknown class, Table~\ref{table:office_W} compares the accuracy when the unknown class is not included in $\mathcal{C}$.  Adding the unknown class improves the accuracy slightly since it enforces that the negative mean of the source is mapped to a negative sample in the target. The impact, however, is very small. 

Additionally, we also analyse the impact of increasing the amount of unknown samples in both source and target domain on the configuration \emph{Amazon} $\rightarrow$ \emph{DSLR+Webcam}. Since the domain shift between \emph{DSLR} and \emph{Webcam} is close to zero (same scenario, but different cameras), they can be merged to get more unknown samples. Following the described protocol, we take 20 samples per known category, also in this case for the target domain, and we randomly increase the number of unknown samples from 20 to 400 in both domains at the same time. As shown in Table~\ref{table:office_num_unknown}, that reports the mean accuracies of 5 random splits, adding more unknown samples decreases the accuracy if domain adaptation is not used (LSVM), but also for the domain adaptation method CORAL~\cite{CNN-DA_Sun15}. This is expected since the unknowns are from different classes and the impact of the unknowns compared to the samples from the shared classes increases. Our method handles such an increase and the accuracies remain stable between 80.3\% and 82.5\%.

\begin{table}[h]
	\scriptsize
	\begin{center}
		\setlength{\tabcolsep}{.375em}
		\begin{tabular}{|l|c|c|c|c|c|c|c|c|}
			\cline{1-9}
			\multicolumn{9}{|c|}{\emph{Amazon} $\rightarrow$ \emph{DSLR+Webcam}} \\ \hline
			\emph{number of unknowns} & 20 & 40 & 60 & 80 & 100 & 200 & 300 & 400 \\ \cline{1-9}
			\emph{unknown / known} & 0.10 & 0.20 & 0.30 & 0.40 & 0.50 & 1.00 & 1.50 & 2.00 \\ \hhline{|=|=|=|=|=|=|=|=|=|}
			LSVM & 74.2 & 70.0 & 66.2 & 63.4 & 61.4 & 53.9 & 50.4 & 48.2 \\ \hline
			CORAL~\cite{CNN-DA_Sun15} & 77.2 & 76.4 & 76.2 & 74.8 & 73.7 & 71.5 & 70.8 & 69.7 \\ \hline
			ATI-$\lambda$ & 80.3 & 82.4 & 81.2 & 81.7 & 82.5 & 80.9 & 80.7 & 81.9 \\ \hline
		\end{tabular}
	\end{center}
		\caption{Impact of increasing the amount of unknown samples in the domain shift \emph{Amazon} $\rightarrow$ \emph{DSLR+Webcam} on the unsupervised Office dataset with 10 shared classes using 20 random samples per known class in both domains.}
		\label{table:office_num_unknown}
\end{table}

\begin{figure*}[t]
	\centering
	\subfigure{
		\includegraphics[width=0.3\linewidth]{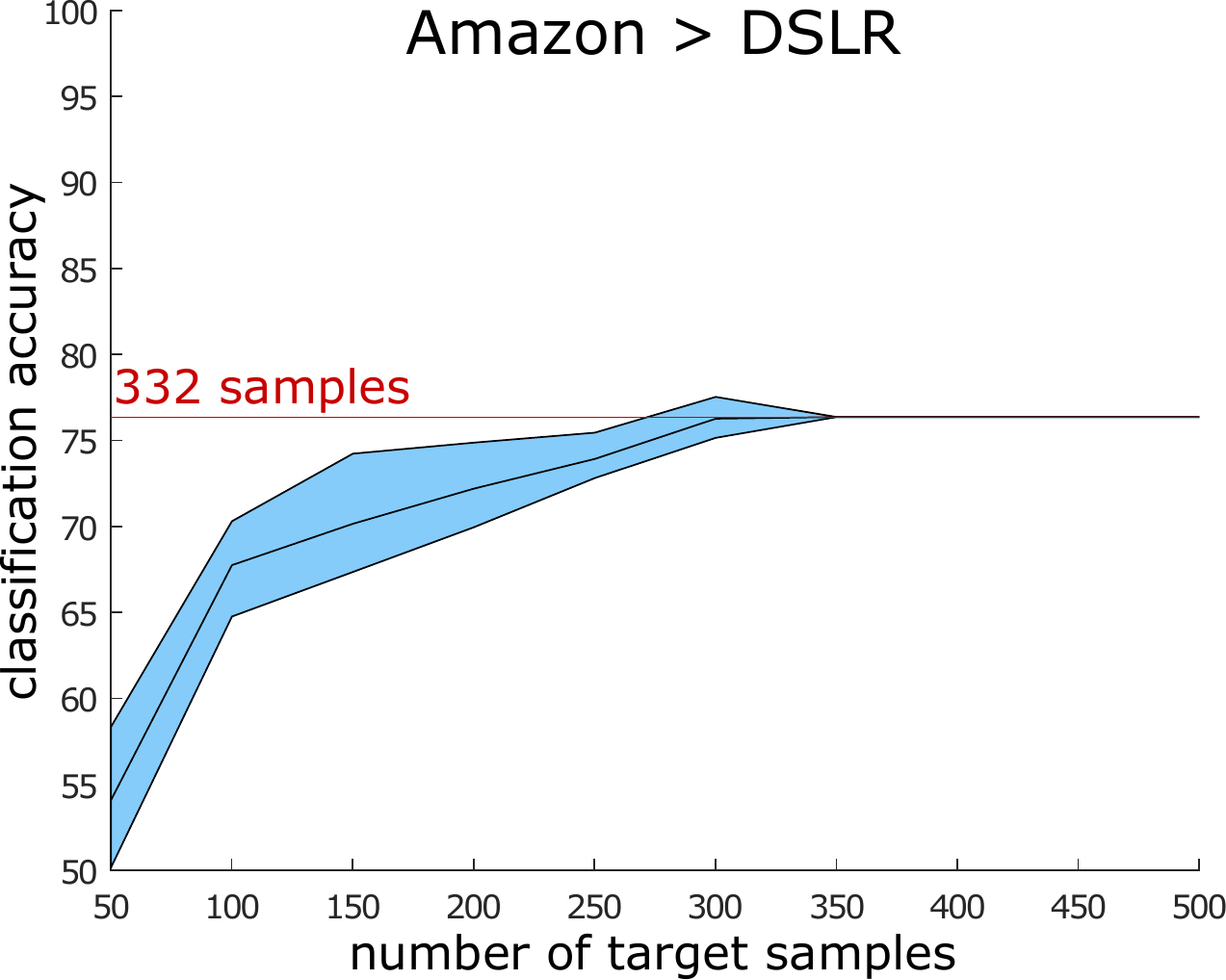}
		\label{fig:numtgt_AD}
	}
	\subfigure{
		\includegraphics[width=0.3\linewidth]{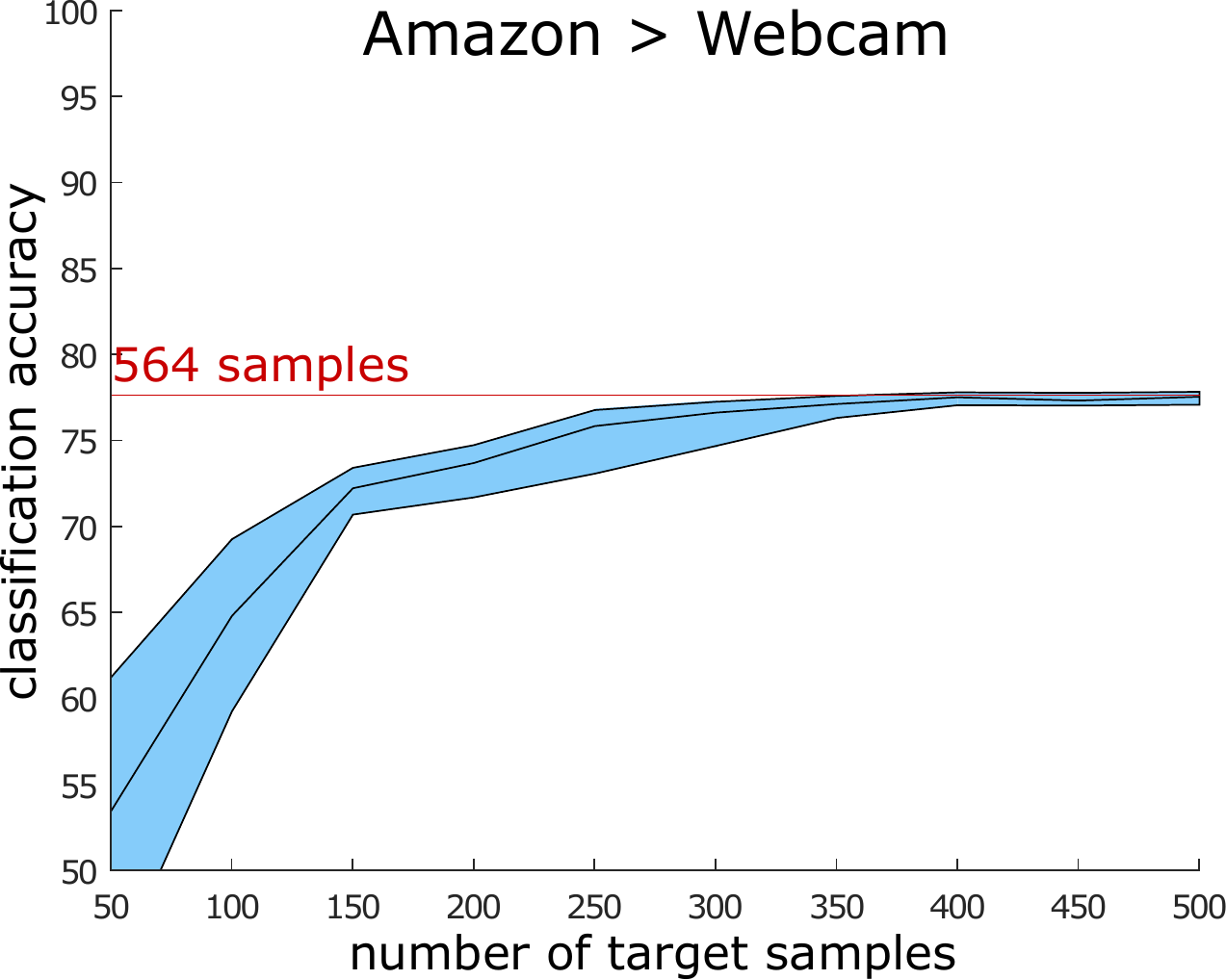}
		\label{fig:numtgt_AW}
	}
	\subfigure{
		\includegraphics[width=0.3\linewidth]{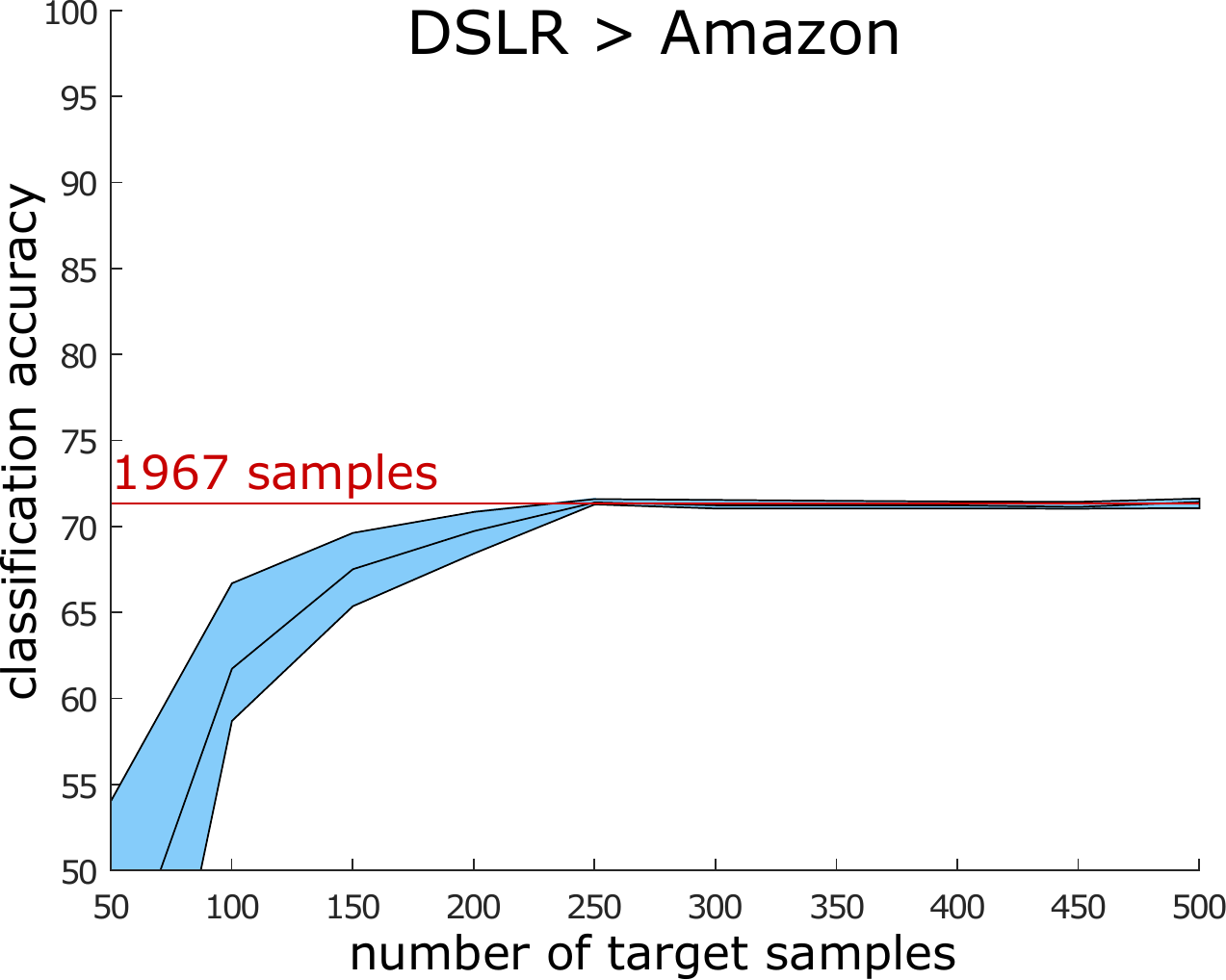}
		\label{fig:numtgt_DA}
	}
	\subfigure{
		\includegraphics[width=0.3\linewidth]{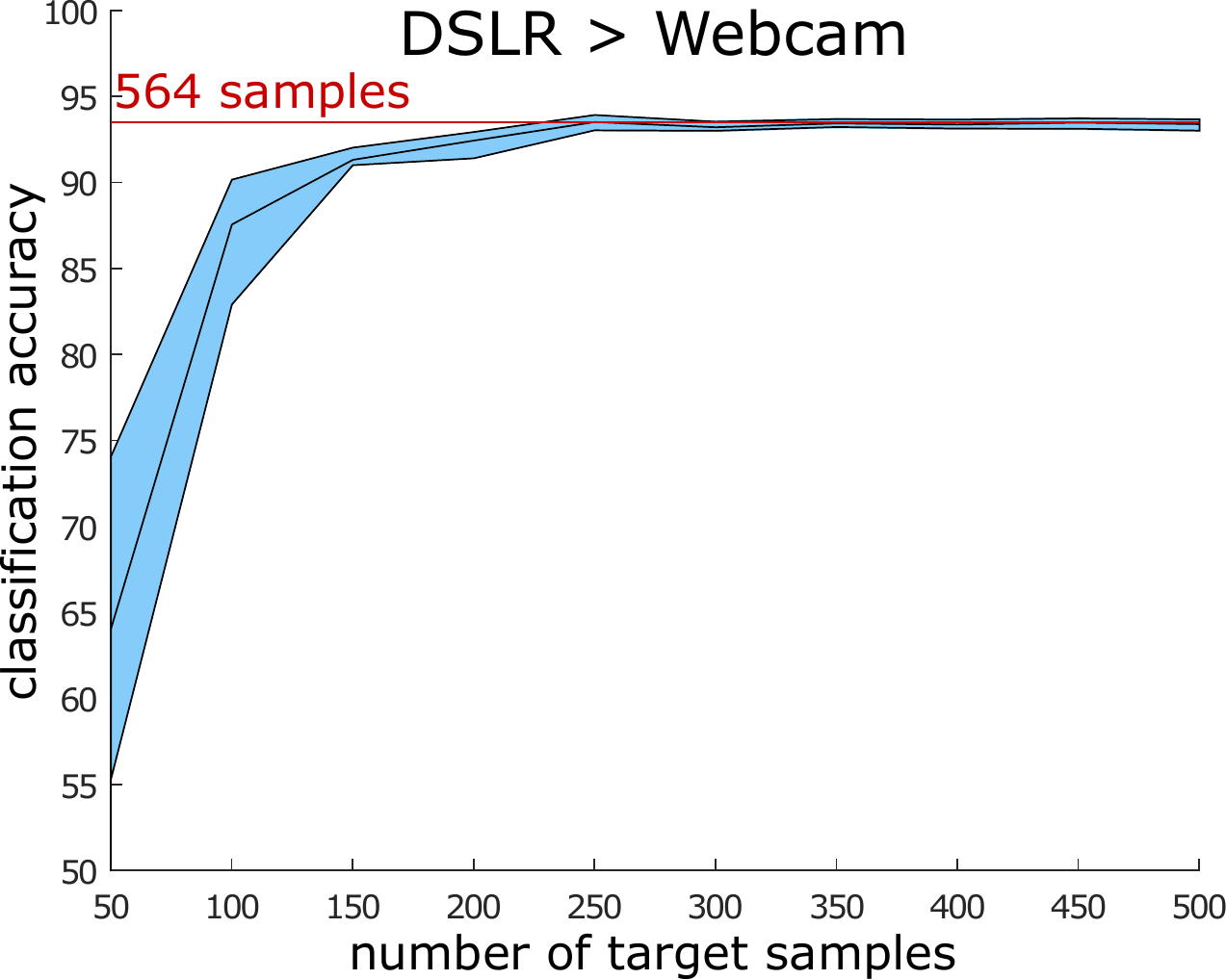}
		\label{fig:numtgt_DW}
	}
	\subfigure{
		\includegraphics[width=0.3\linewidth]{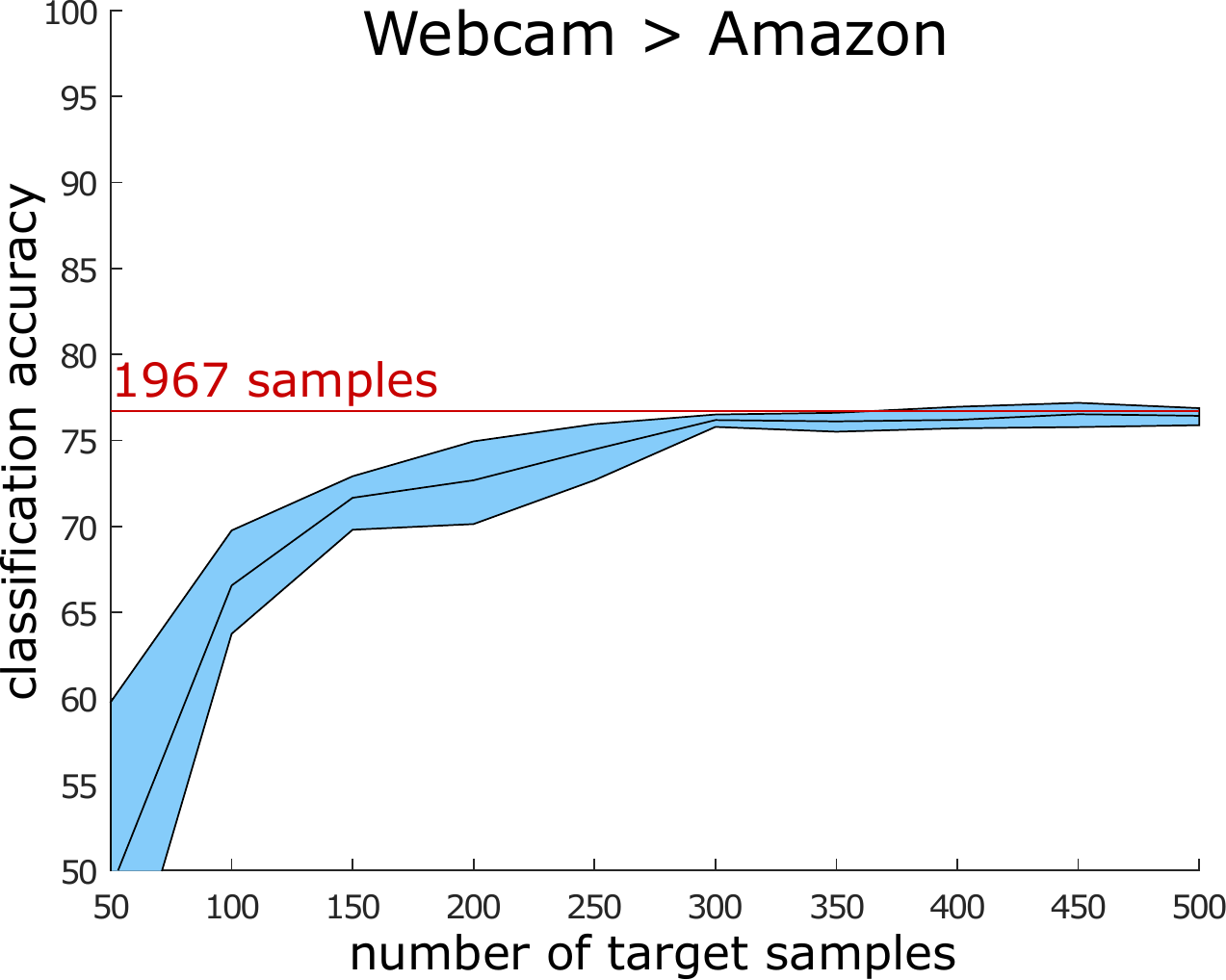}
		\label{fig:numtgt_WA}
	}
	\subfigure{
		\includegraphics[width=0.3\linewidth]{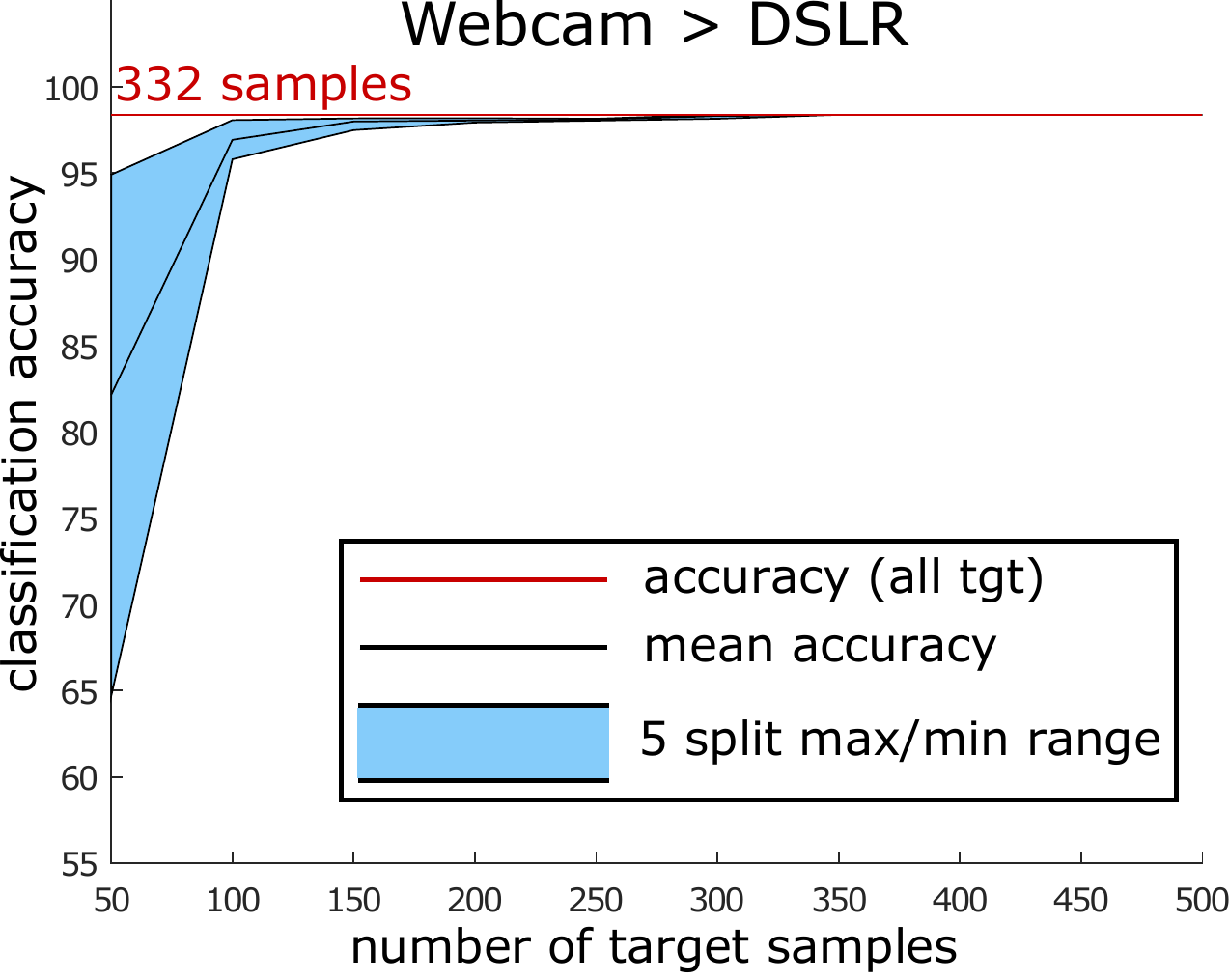}
		\label{fig:numtgt_WD}
	}
	\caption{Impact of using a random subset of target samples. The blue region shows the difference between the best and worst result of the 5 randomly sampled subsets for a given number of target samples and the black line within the region is the mean accuracy of the 5 subsets.     
	The red line indicates the classification accuracy when using all target samples. The results are reported for ATI-$\lambda$ using the open set protocol on the unsupervised Office dataset with 10 shared classes using all samples per class.
	}
	\label{fig:numtgt}
\end{figure*}

\vspace{1mm}\noindent{\bf Subsampling of target samples.}
In order to evaluate the robustness of our method when having a reduced amount of target samples for domain adaptation, we subsample the target data. Fig.~\ref{fig:numtgt} shows the results for ATI-$\lambda$ on the 6 domain shifts of the Office dataset with the standard open set protocol (OS). We vary the number of target samples from 50 to the total number of instances. For a fixed number of target samples, we randomly sample 5 times from the target data and plot the lowest, highest and average accuracy of the 5 runs. The accuracy is always measured on the whole target dataset.      
The results show that between 300 and 400 target instances are sufficient to achieve similar accuracies than our method with all target samples. 
When the domain shifts are smaller, \eg, \emph{D$\rightarrow$ W} and \emph{W$\rightarrow$ D}, even less target samples are required.

\begin{figure}[h]
	\centering
	\subfigure{
		\includegraphics[width=0.9\linewidth]{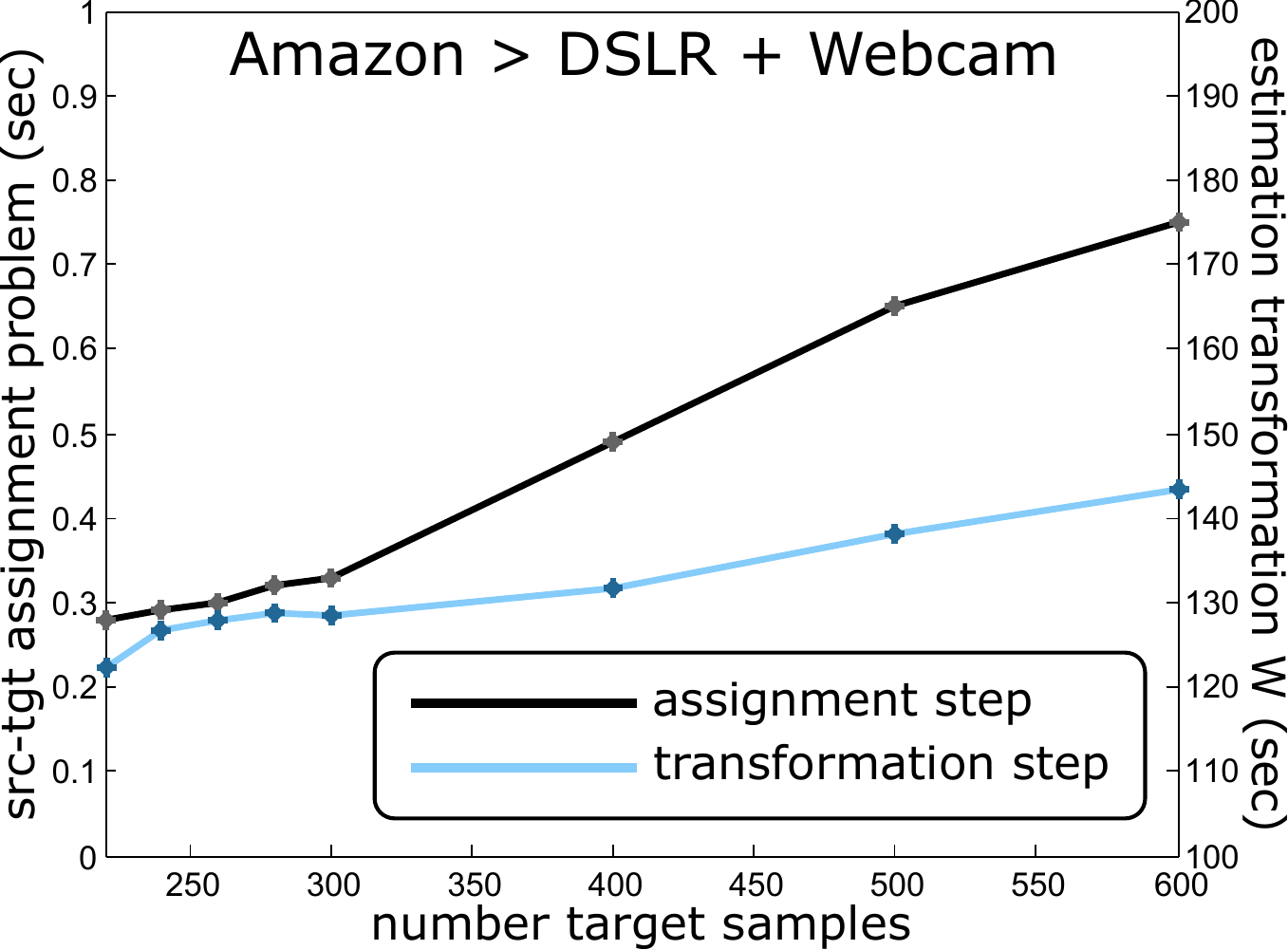}
	}
	\caption{Execution time in seconds for the assignment and transformation estimation steps of a single iteration with respect to the number of target samples. 
	}
	\label{fig:timings}
\end{figure}

{
\vspace{1mm}\noindent{\bf Scalability analysis of target samples.} The number of sampled target samples has an impact on the execution time of the assignment and the transformation steps of the iterative process.
Therefore, we also test the scalability of the two steps of our method with respect to the number of target samples.
The average execution times of both techniques in the domain shift \emph{Amazon} $\rightarrow$ \emph{DSLR+Webcam} for all the random splits and unknown sets of the previous evaluation are shown in Fig.~\ref{fig:timings}. We observe that the assignment problem takes less than a second to be solved for any size of target data from the evaluated settings. Most of the computation time is required for estimating the transformation $W$, which requires at least 120 seconds. The computation time of this step, however, increases only moderately with respect to the number of target samples.   
}

\begin{figure*}[t]
	\centering
	\subfigure{
		\includegraphics[width=0.3\linewidth]{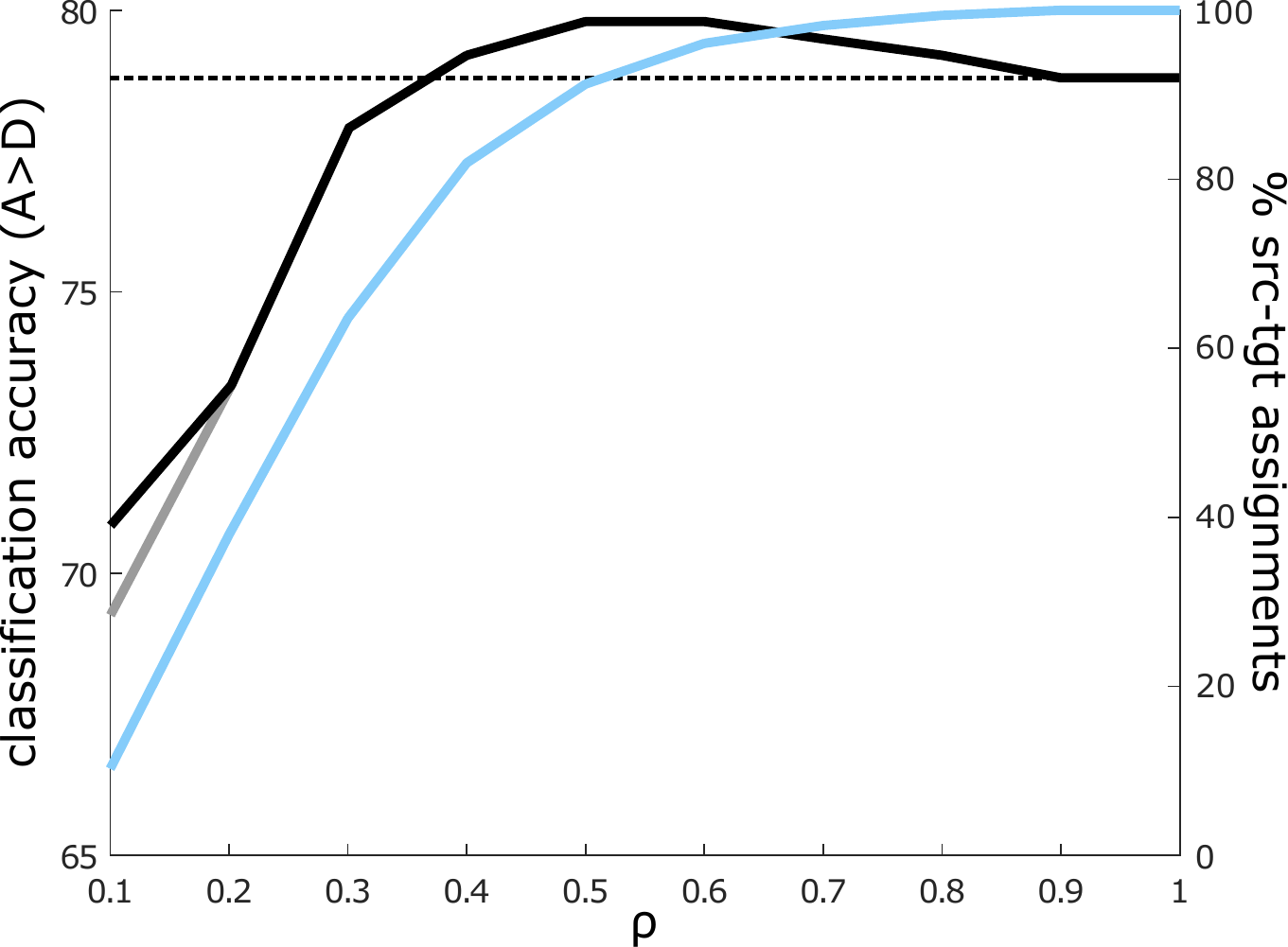}
		\label{fig:lambda_AD}
	}
	\subfigure{
		\includegraphics[width=0.3\linewidth]{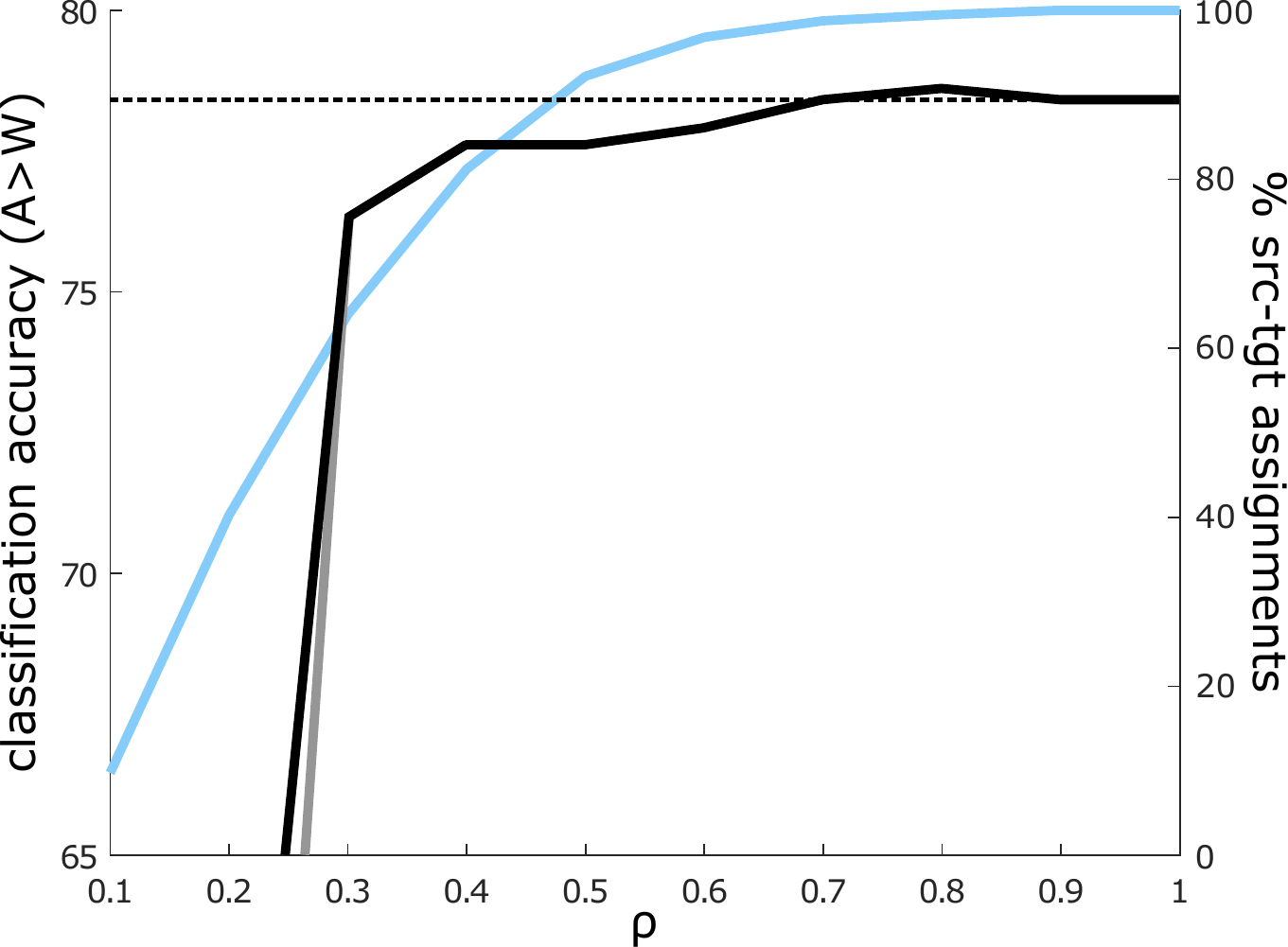}
		\label{fig:lambda_AW}
	}
	\subfigure{
		\includegraphics[width=0.3\linewidth]{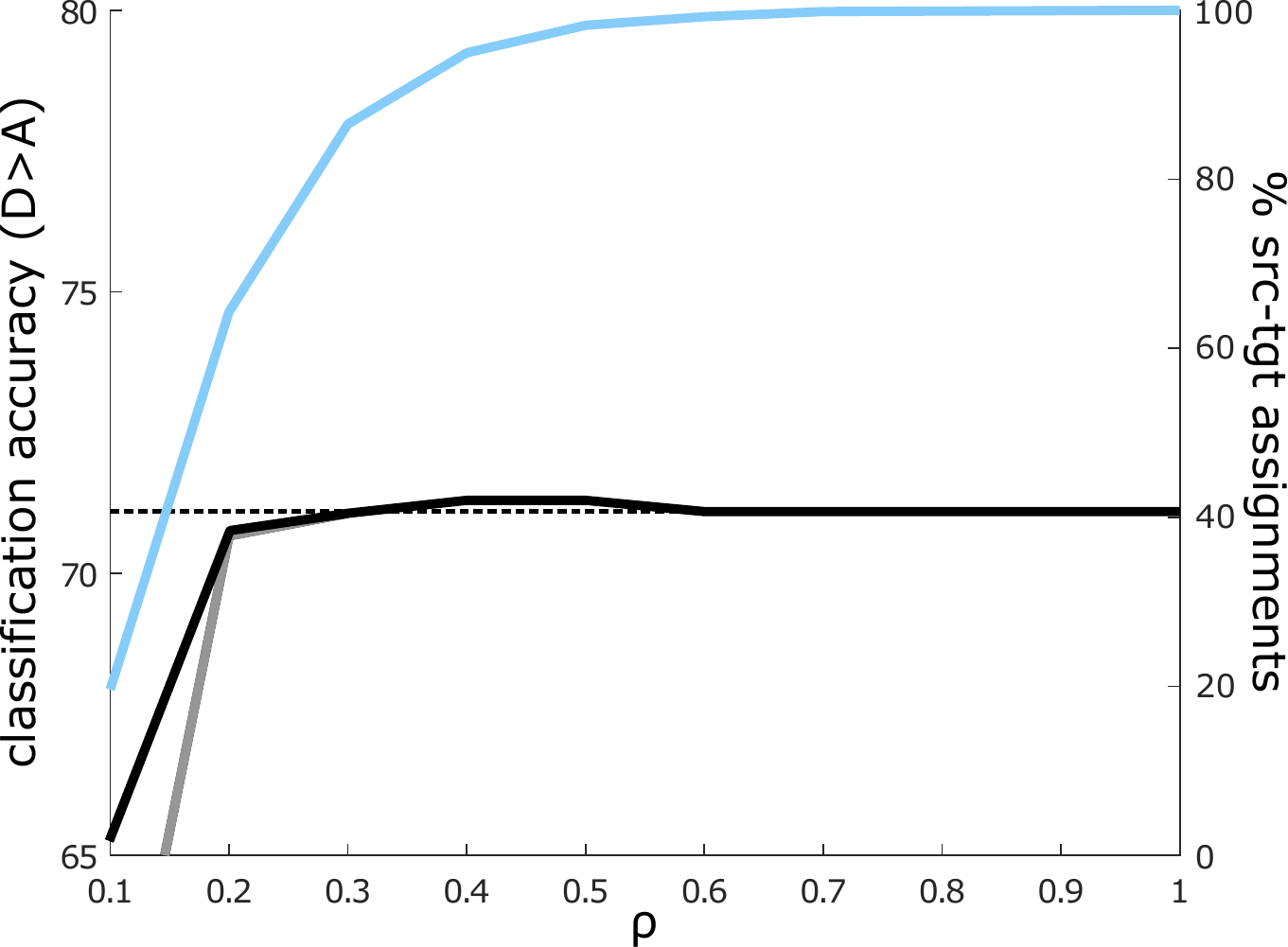}
		\label{fig:lambda_DA}
	}
	\subfigure{
		\includegraphics[width=0.3\linewidth]{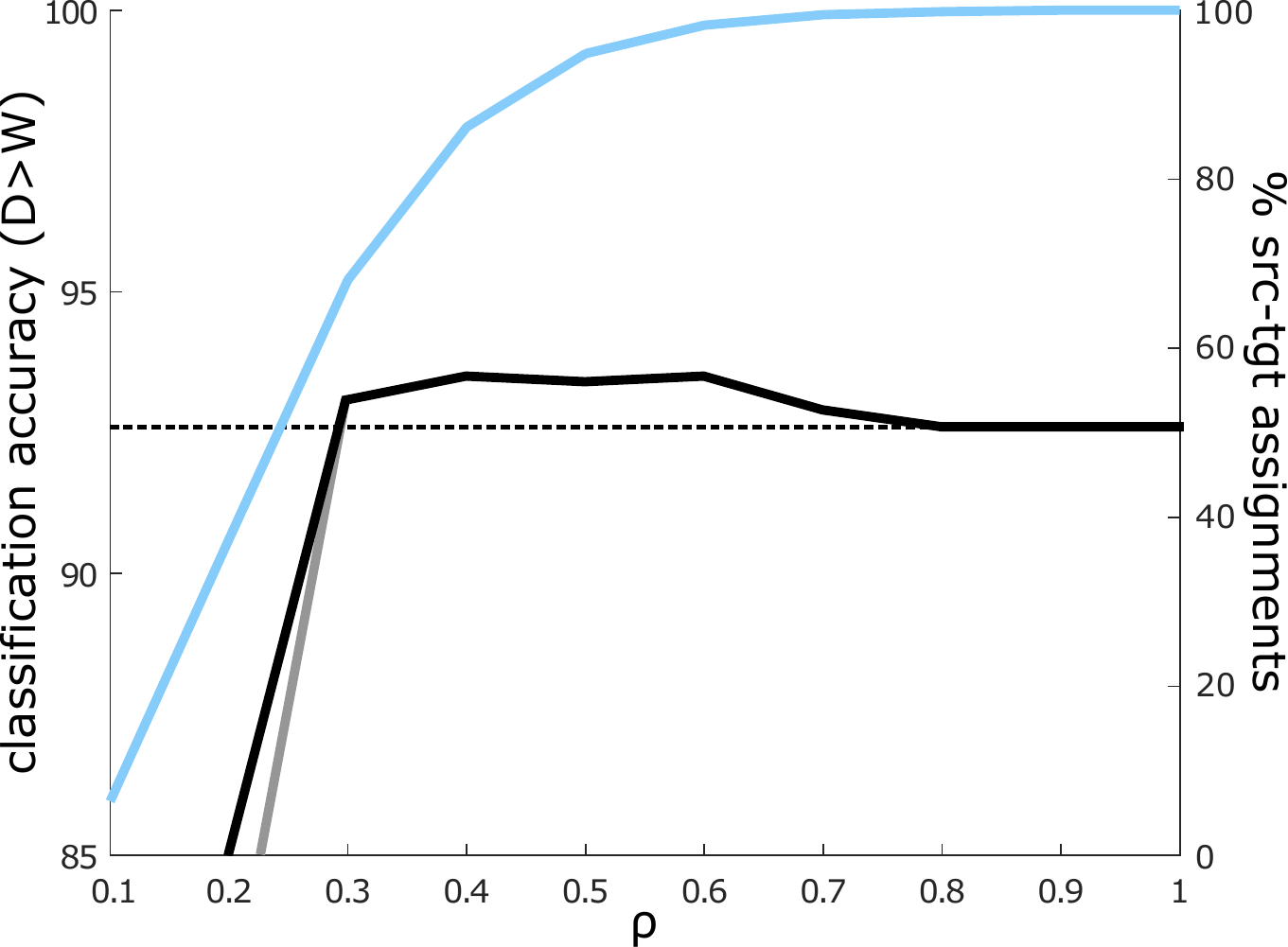}
		\label{fig:lambda_DW}
	}
	\subfigure{
		\includegraphics[width=0.3\linewidth]{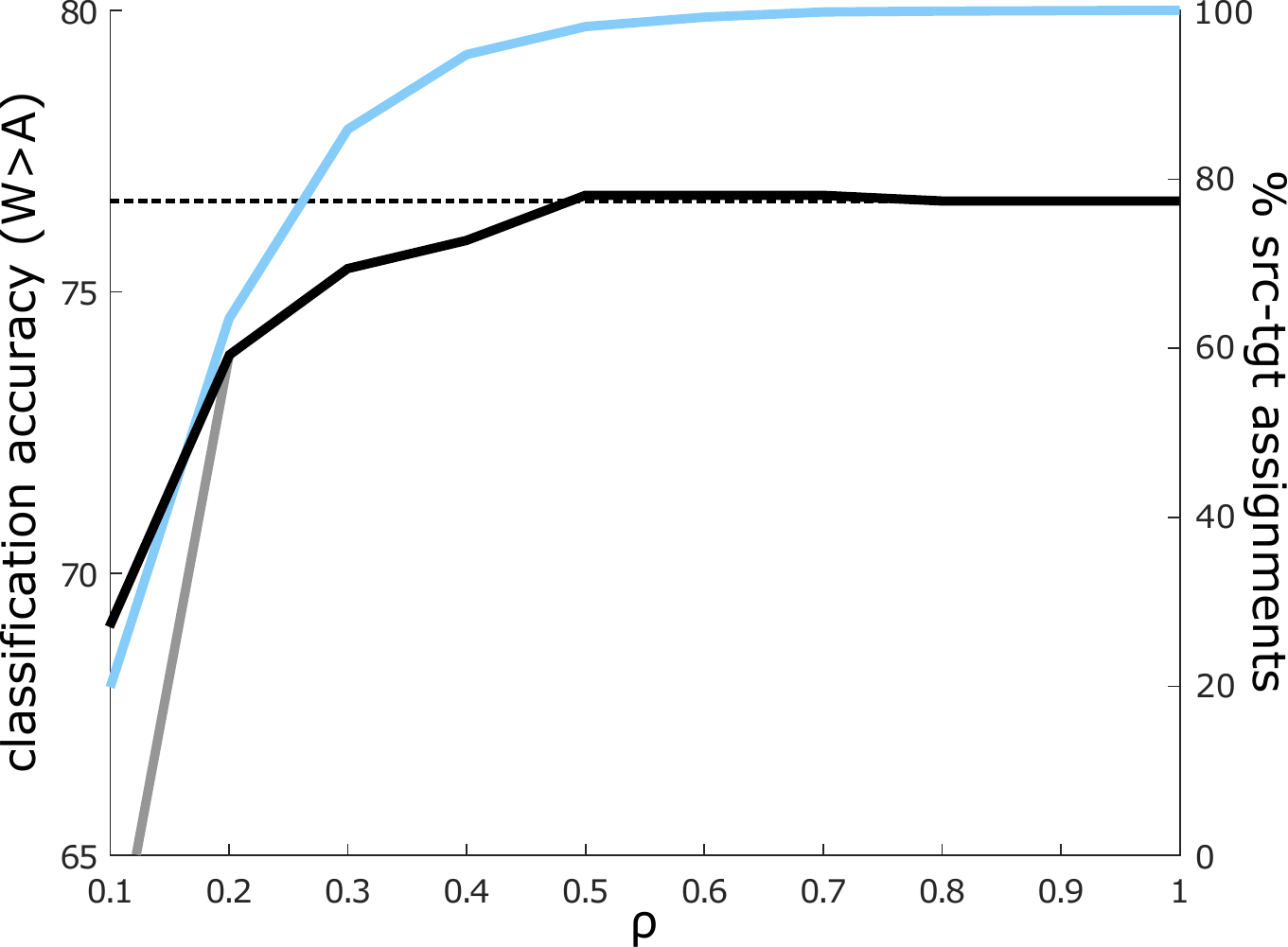}
		\label{fig:lambda_WA}
	}
	\subfigure{
		\includegraphics[width=0.3\linewidth]{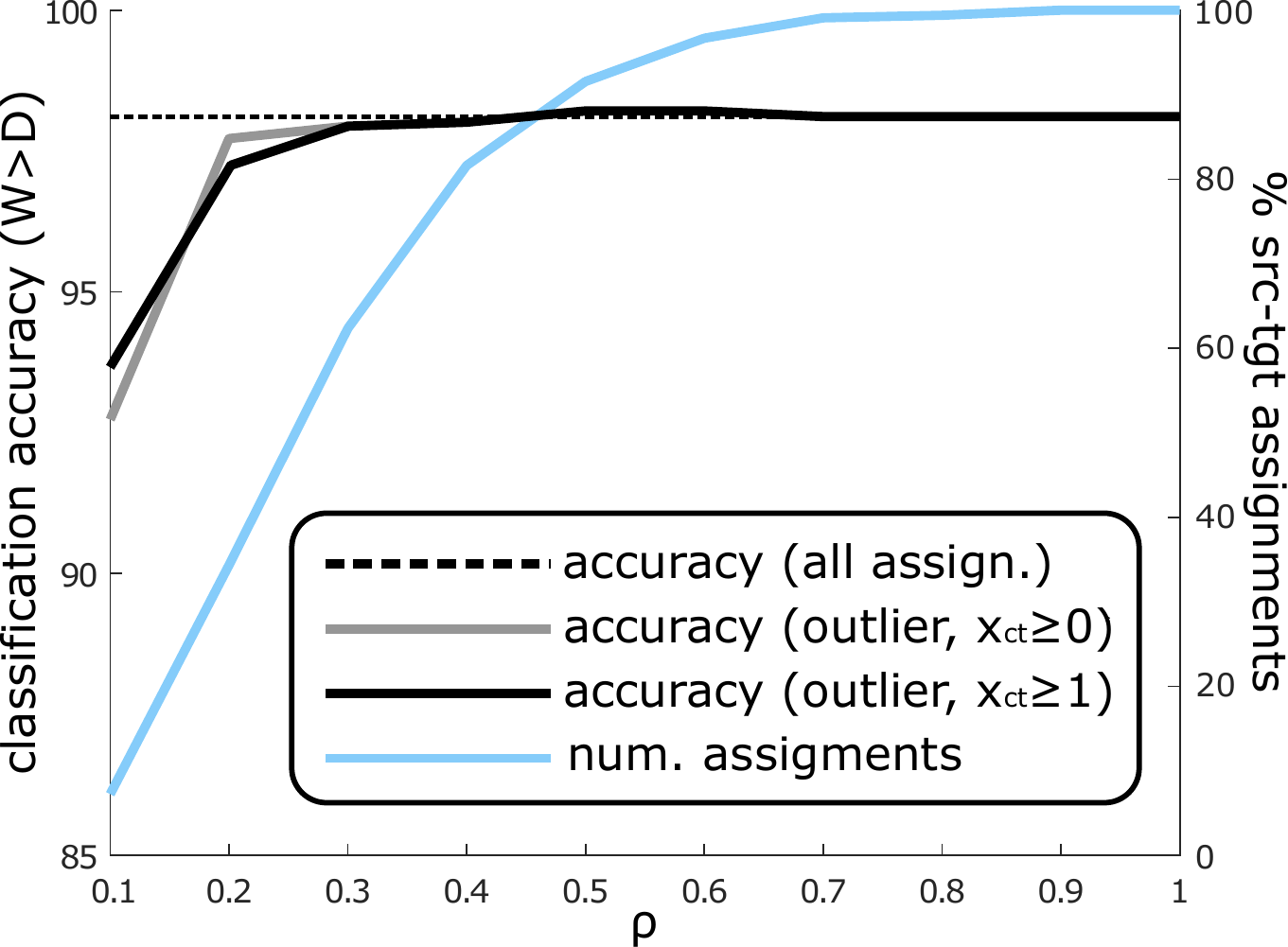}
		\label{fig:lambda_WD}
	}
	\caption{The black and grey curves show the classification accuracies for varying values of $\rho$ when including or not the constraint $\sum_{t} x_{ct} \geq 1$, respectively. 
	$\rho = 0.5$ obtains the best accuracies in 5 out of 6 domain shifts. The blue curve shows the percentage of selected assignments to compute the transformation matrix $W$ in the first iteration. The results are reported for ATI-$\lambda$ using the open set protocol on the unsupervised Office dataset with 10 shared classes using all samples per class.}
	\label{fig:lambda}
\end{figure*}

\vspace{1mm}\noindent{\bf Impact of parameter $\rho$.}
The cost that determines whether a target sample is considered as outlier during the assignment process is defined by $\lambda$ \eqref{eq:lambda}, which is based on the current minimum and maximum distance between the source clusters and target samples. Thus, $\lambda$ is updated at each iteration. The value of $\lambda$, however, also depends on the parameter $\rho$. 
For all experiments, we use $\rho = 0.5$ as default value, aiming for a moderate outlier rejection. 
Fig.~\ref{fig:lambda} shows the impact of $\rho$ on the accuracy. 
Using  $\rho = 0.5$, which rejects around 10-20$\%$ of the target samples, achieves the best results in 5 out of the 6 domain shifts on the Office dataset.
When $\rho$ gets closer to 0 the accuracy drops substantially since too many samples are discarded. 

{
\vspace{1mm}\noindent{\bf Impact of constraint $\sum_{t} x_{ct} \geq 1$.}
Our formulation in~(\ref{eq:assignments}) ensures that at least one target sample is assigned to an object category. Therefore, all classes contribute to the estimation of the transformation matrix $W$. In order to measure its impact on the adaptation problem, we run experiments with $\sum_{t} x_{ct} \geq 1$ and without the constraint, i.e., when a class might not be assigned to any target sample at all. As illustrated in Fig.~\ref{fig:lambda}, the inclusion of this constraint provides higher accuracies when $\rho < 0.3$. For greater values of $\rho$, the constraint can be omitted since it does not influence the accuracy.
}

{
\vspace{1mm}\noindent{\bf Impact of wrong assignments.} 
During the iterative process of our method, wrong assignments take part in the optimisation of $W$, introducing false associations between the source and the target domain that negatively affect the final transformation. A general assumption in our method is that the correct assignments largely compensate the wrong ones and, thus, the transformed source data allows for better classification accuracies in the target domain. Therefore, we artificially generate assignments in the first iteration by assigning a random subset of target samples to the correct class in the source domain and the remaining target samples to random classes. We then run our approach without any additional modifications until it converges.    
We report in Table~\ref{table:gt_rnd_evolution} the average percentage of correct assignments of 5 random splits for the domain shift \emph{Amazon} $\rightarrow$ \emph{DSLR+Webcam} with 400 unknown samples.
While the first iteration represents the accuracy of correct and random assignments that we generate, the last row shows the accuracies after the approach has converged. 
As it can be observed, the approach ends in a local optimum, but the accuracies increase for all cases except if we initialise the approach with 100\% correct assignments. It is expected that the assignment accuracy does not remain at 100\% since the image manifolds are not perfectly linearised and even for the best estimate of $W$ wrong assignments can occur.            
}
\begin{table}[h]
	\scriptsize
	\begin{center}
		\setlength{\tabcolsep}{.375em}
		\begin{tabular}{|c|c|c|c|c|c|c|c|c|c|c||c|}
			\cline{1-12}
			\multicolumn{12}{|c|}{\emph{Amazon} $\rightarrow$ \emph{DSLR+Webcam} (400 unknown samples)} \\ \hline
			$\%$gt (+rnd) & 10 & 20 & 30 & 40 & 50 & 60 & 70 & 80 & 90 & 100 & \emph{std} \\ \hline
			iteration 1 & 18.2 & 27.0 & 36.1 & 45.2 & 54.3 & 63.5 & 72.7 & 81.7  & 90.7 & 100.0 & \emph{85.1} \\ \hline
			final & 24.4 & 40.1 & 54.7 & 65.4 & 72.8 & 79.2 & 83.6 & 88.8 & 93.1 & 96.7 & \emph{88.6} \\ \hline
		\end{tabular}
	\end{center}
	\caption{Impact of limiting the amount of correct assignments in the first iteration. We report the average percentage of correct assignments over 5 random splits and increase the percentage of correctly selected assignments from 10$\%$ to 100$\%$, leaving the rest randomly selected. The last column shows the percentage of correct assignments of the method without modifying the initial assignments.}
	\label{table:gt_rnd_evolution}
\end{table}

\vspace{1mm}\noindent{\bf Semi-supervised domain adaptation.}
We also evaluate our approach for open set domain adaptation on the \emph{Office} dataset in its semi-supervised setting. Applying again the standard protocol of~\cite{DA_Saenko10} with the subset of source samples, we also take 3 labelled target samples per class and leave the rest unlabelled. We compare our method with the deep learning method MMD~\cite{CNN-DA_Tzeng14}.
As baselines, we report the accuracy for the linear SVMs without domain adaptation (LSVM) when they are trained only on the source samples (s), only on the annotated target samples (t) or on both (st). 
As expected, the baseline trained on both performs best as shown in Table~\ref{table:office_sup}. Our approach ATI outperforms the baseline and the CNN approach~\cite{CNN-DA_Tzeng14}. As in the unsupervised case, the improvement compared to the CNN approach is larger for the open set (+4.8\%) than for the closed set (+2.2\%). While the locality constrained formulation, ATI-$\lambda$-$N$, decreased the accuracy for the unsupervised setting, it improves the accuracy for the semi-supervised case since the formulation enforces that neighbours of the target samples are assigned to the same class. The results with one (ATI-$\lambda$-$N1$) or two neighbours (ATI-$\lambda$-$N2$) are similar.  

\begin{table}[h]
	\scriptsize
	\begin{center}
		\setlength{\tabcolsep}{0.14em}
		\begin{tabular}{|l|c|c|c|c|c|c|}
			\cline{2-7}
			\multicolumn{1}{c|}{} &
			\multicolumn{3}{c|}{A$\rightarrow$D} & \multicolumn{3}{c|}{A$\rightarrow$W} \\ \cline{2-7}
			\multicolumn{1}{c|}{} &
			CS (10) & OS$^*$ (10) & OS (10) & CS (10) & OS$^*$ (10) & OS (10) \\ \hhline{|=|=|=|=|=|=|=|}
			LSVM (s) & 85.8$\pm$3.2 & 62.1$\pm$7.9 & 65.9$\pm$6.2 & 76.4$\pm$2.1 & 45.7$\pm$5.0 & 50.4$\pm$4.5 \\ \hline
			LSVM (t) & 92.3$\pm$3.9 & 68.2$\pm$5.2 & 71.1$\pm$4.7 & 91.5$\pm$4.9 & 59.6$\pm$3.7 & 63.2$\pm$3.4 \\ \hline
			LSVM (st) & 95.7$\pm$1.3 & 82.5$\pm$3.0 & 84.0$\pm$2.6 & 92.4$\pm$1.8 & 72.5$\pm$3.7 & 74.8$\pm$3.4 \\ \hhline{|=|=|=|=|=|=|=|}				
			MMD~\cite{CNN-DA_Tzeng14} & 94.1$\pm$2.3 & 86.1$\pm$2.3 & 86.8$\pm$2.2 & 92.4$\pm$2.8 & 76.4$\pm$1.5 & 78.3$\pm$1.3 \\ \hhline{|=|=|=|=|=|=|=|}
			ATI & 95.4$\pm$1.3 & 89.0$\pm$1.4 & 89.7$\pm$1.3 & 95.9$\pm$1.3 & 84.0$\pm$1.7 & 85.1$\pm$1.5 \\ \hline
			ATI-$\lambda$ & 97.1$\pm$1.1 & \textbf{89.5$\pm$1.4} & 90.2$\pm$1.3 & 96.1$\pm$2.0 & 84.1$\pm$1.8 & 85.2$\pm$1.5 \\ \hline
			ATI-$\lambda$-N1 & 97.6$\pm$1.0 & \textbf{89.5$\pm$1.3} & \textbf{90.3$\pm$1.2} & \textbf{96.4$\pm$1.7} & \textbf{84.4$\pm$3.6} & \textbf{85.5$\pm$1.5} \\ \hline
			ATI-$\lambda$-N2 & \textbf{97.9$\pm$1.4} & 89.4$\pm$1.2 & 90.1$\pm$1.0 & 92.8$\pm$1.6 & 84.3$\pm$2.4 & 85.4$\pm$1.5 \\ \hline			
		\end{tabular}
		\begin{tabular}{|l|c|c|c|c|c|c|}
			\cline{2-7}
			\multicolumn{1}{c|}{} &
			\multicolumn{3}{c|}{D$\rightarrow$A} & \multicolumn{3}{c|}{D$\rightarrow$W} \\ \cline{2-7}
			\multicolumn{1}{c|}{} &
			CS (10) & OS$^*$ (10) & OS (10) & CS (10) & OS$^*$ (10) & OS (10) \\ \hhline{|=|=|=|=|=|=|=|}
			LSVM (s) & 85.2$\pm$1.7 & 40.3$\pm$4.3 & 45.2$\pm$3.8 & 97.2$\pm$0.7 & 81.4$\pm$2.4 & 83.0$\pm$2.2 \\ \hline
			LSVM (t) & 88.7$\pm$2.2 & 52.8$\pm$6.0 & 57.0$\pm$5.5 & 91.5$\pm$4.9 & 59.6$\pm$3.7 & 63.2$\pm$3.4 \\ \hline
			LSVM (st) & 91.9$\pm$0.7 & 68.7$\pm$2.5 & 71.2$\pm$2.3 & 98.7$\pm$0.9 & 87.3$\pm$2.3 & 88.5$\pm$2.1 \\ \hhline{|=|=|=|=|=|=|=|}	
			MMD~\cite{CNN-DA_Tzeng14} & 90.2$\pm$1.8 & 69.0$\pm$3.4 & 71.3$\pm$3.0 & 98.5$\pm$1.0 & 85.5$\pm$1.6 & 86.7$\pm$1.4 \\ \hhline{|=|=|=|=|=|=|=|}
			ATI & \textbf{93.5$\pm$0.2} & 74.4$\pm$2.7 & 76.1$\pm$2.5 & 98.7$\pm$0.7 & 91.6$\pm$1.7 & 92.4$\pm$1.5 \\ \hline
			ATI-$\lambda$ & \textbf{93.5$\pm$0.2} & 74.4$\pm$2.5 & 76.2$\pm$2.3 & 98.7$\pm$0.8 & 91.6$\pm$1.7 & 92.4$\pm$1.5 \\ \hline
			ATI-$\lambda$-N1 & 93.4$\pm$0.2 & 74.6$\pm$2.5 & 76.4$\pm$2.3 & 98.9$\pm$0.5 & 92.0$\pm$1.6 & 92.7$\pm$1.5 \\ \hline
			ATI-$\lambda$-N2 & \textbf{93.5$\pm$0.1} & \textbf{74.9$\pm$2.3} & \textbf{76.7$\pm$2.1} & \textbf{99.3$\pm$0.5} & \textbf{92.2$\pm$1.9} & \textbf{92.9$\pm$1.7} \\ \hline	
		\end{tabular}
		\begin{tabular}{|l|c|c|c|c|c|c|c|c|c|}
			\cline{2-10}
			\multicolumn{1}{c|}{} &
			\multicolumn{3}{c|}{W$\rightarrow$A} & \multicolumn{3}{c|}{W$\rightarrow$D} & \multicolumn{3}{c|}{AVG.} \\ \cline{2-10}
			\multicolumn{1}{c|}{} & 
			CS (10) & OS$^*$ (10) & OS (10) & CS (10) & OS$^*$ (10) & OS (10) & CS & OS$^*$ & OS \\ \hhline{|=|=|=|=|=|=|=|=|=|=|}
			LSVM (s) & 78.8$\pm$2.9 & 32.4$\pm$3.8 & 38.2$\pm$3.4 & 99.5$\pm$0.3 & 88.7$\pm$2.2 & 89.6$\pm$1.9 & 87.1 & 58.4 & 62.0 \\ \hline
			LSVM (t) & 88.7$\pm$2.2 & 52.8$\pm$6.0 & 57.0$\pm$5.5 & 92.3$\pm$3.9 & 68.2$\pm$5.2 & 71.1$\pm$4.7 & 90.9 & 60.2 & 63.8 \\ \hline
			LSVM (st) & 90.8$\pm$1.3 & 66.2$\pm$4.4 & 69.0$\pm$4.1 & 99.4$\pm$0.7 & 93.5$\pm$2.7 & 94.0$\pm$2.5 & 94.8 & 78.4 & 80.3 \\ \hhline{|=|=|=|=|=|=|=|=|=|=|}	
			MMD~\cite{CNN-DA_Tzeng14} & 89.1$\pm$3.2 & 65.1$\pm$3.8 & 67.8$\pm$3.4 & 98.2$\pm$1.4 & 93.9$\pm$2.9 & 94.4$\pm$2.7 & 93.8 & 79.3 & 80.9 \\ \hhline{|=|=|=|=|=|=|=|=|=|=|}
			ATI & \textbf{93.0$\pm$0.5} & 71.3$\pm$4.6 & 74.3$\pm$4.3 & 99.3$\pm$0.6 & 96.3$\pm$1.8 & 96.6$\pm$1.7 & 96.0 & 84.4 & 85.7 \\ \hline
			ATI-$\lambda$ & \textbf{93.0$\pm$0.5} & 71.5$\pm$4.8 & 73.6$\pm$4.4 & \textbf{99.5$\pm$0.6} & 96.3$\pm$1.8 & 96.6$\pm$1.7 & 96.3 & 84.6 & 85.7 \\ \hline
			ATI-$\lambda$-N1 & \textbf{93.0$\pm$0.6} & 72.2$\pm$4.5 & 74.2$\pm$4.1 & 99.3$\pm$0.6 & \textbf{96.7$\pm$2.1} & \textbf{97.0$\pm$1.9} & 96.4 & \textbf{84.9} & \textbf{86.0} \\ \hline
			ATI-$\lambda$-N2 & \textbf{93.0$\pm$0.6} & \textbf{72.8$\pm$4.2} & \textbf{74.8$\pm$3.9} & 99.3$\pm$0.6 & 95.5$\pm$2.2 & 95.9$\pm$2.0 & \textbf{96.6} & 84.8 & \textbf{86.0}  \\ \hline			
		\end{tabular}
		\end{center}
		\vspace{2mm}	
		\caption{Open set domain adaptation on the semi-supervised Office dataset with 10 shared classes (OS). We report the average and the standard deviation using a subset of samples per class in 5 random splits~\cite{DA_Saenko10}.}
		\label{table:office_sup}
\end{table}

\subsubsection{Dense Cross-Dataset Analysis} 

\begin{table*}[t]
	\scriptsize
	\begin{center}
		\setlength{\tabcolsep}{0.2em}
		\begin{tabular}{|l|c|c|c|c|c|c|c|c|c|c|c|c|}
			\cline{2-13}
			\multicolumn{1}{c|}{} &
			\multicolumn{2}{c|}{B$\rightarrow$C} & \multicolumn{2}{c|}{B$\rightarrow$I} & \multicolumn{2}{c|}{B$\rightarrow$S} & \multicolumn{2}{c|}{C$\rightarrow$B} & \multicolumn{2}{c|}{C$\rightarrow$I} & \multicolumn{2}{c|}{C$\rightarrow$S} \\ \cline{2-13}
			\multicolumn{1}{c|}{} &
			CS (10) & OS (10) & CS (10) & OS (10) & CS (10) & OS (10) & CS (10) & OS (10) & CS (10) & OS (10) & CS (10) & OS (10) \\ \hhline{|=|=|=|=|=|=|=|=|=|=|=|=|=|}
			LSVM & 82.4$\pm$2.4 & 66.6$\pm$4.0 & 75.1$\pm$0.4 & 59.0$\pm$2.7 & 43.0$\pm$2.0 & 24.2$\pm$3.0 & 53.5$\pm$2.1 & 40.1$\pm$1.9 & 76.9$\pm$4.3 & 62.5$\pm$1.2 & 46.3$\pm$2.7 & 28.2$\pm$1.4 \\ \hhline{|=|=|=|=|=|=|=|=|=|=|=|=|=|}
			TCA~\cite{DA_Pan09} & 74.9$\pm$3.0 & 62.8$\pm$3.8 & 68.4$\pm$4.0 & 56.6$\pm$4.5 & 38.3$\pm$1.7 & 29.6$\pm$4.2 & 49.2$\pm$1.1 & 38.9$\pm$1.9 & 73.1$\pm$3.6 & 60.2$\pm$1.4 & 45.9$\pm$3.6 & 29.7$\pm$1.6 \\ \hline
			gfk~\cite{DA_Gong12} & 82.0$\pm$2.2 & 66.2$\pm$4.0 & 74.3$\pm$1.0 & 58.3$\pm$3.1 & 42.2$\pm$1.4 & 23.8$\pm$2.0 & 53.2$\pm$2.6 & 40.2$\pm$1.8 & 77.1$\pm$3.3 & 62.2$\pm$1.5 & 46.2$\pm$3.0 & 28.5$\pm$1.0 \\ \hline
			SA~\cite{DA_Fernando13} & 81.1$\pm$1.8 & 66.0$\pm$3.4 & 73.9$\pm$0.9 & 57.8$\pm$3.2 & 41.9$\pm$2.4 & 24.3$\pm$2.6 & 53.4$\pm$2.5 & 40.3$\pm$1.7 & 77.3$\pm$4.2 & 62.5$\pm$.8 & 46.1$\pm$3.3 & 29.0$\pm$1.5 \\ \hline
			CORAL~\cite{CNN-DA_Sun15} & 80.1$\pm$3.5 & 68.8$\pm$3.3 & 73.7$\pm$2.0 & 60.9$\pm$2.6 & 42.2$\pm$2.4 & 27.2$\pm$3.9 & 53.6$\pm$2.9 & 40.7$\pm$1.5 & 78.2$\pm$5.1 & 64.0$\pm$2.6 & 48.2$\pm$3.9 & 31.4$\pm$0.8 \\ \hhline{|=|=|=|=|=|=|=|=|=|=|=|=|=|}
			ATI & 86.3$\pm$1.6 & \textbf{71.4$\pm$1.8} & 80.1$\pm$0.7 & 68.0$\pm$1.9 & \textbf{49.2$\pm$3.2} & 36.8$\pm$1.2 & 53.2$\pm$3.4 & 45.4$\pm$3.4 & 81.7$\pm$3.7 & 66.7$\pm$4.2 & 52.0$\pm$3.4 & 35.8$\pm$1.8 \\ \hline
			ATI-$\lambda$ & \textbf{86.7$\pm$1.3} & \textbf{71.4$\pm$2.3} & \textbf{80.6$\pm$2.4} & \textbf{69.0$\pm$2.8} & 48.6$\pm$2.5 & \textbf{37.4$\pm$2.6} & \textbf{54.2$\pm$1.9} & \textbf{45.7$\pm$3.0} & \textbf{82.2$\pm$3.7} & \textbf{67.9$\pm$4.2} & \textbf{53.1$\pm$2.8} & \textbf{37.5$\pm$2.7} \\ \hline 
		\end{tabular}
	\end{center}
	\begin{center}
		\setlength{\tabcolsep}{0.2em}
		\begin{tabular}{|l|c|c|c|c|c|c|c|c|c|c|c|c|c|c|}
			\cline{2-15}
			\multicolumn{1}{c|}{} &
			\multicolumn{2}{c|}{I$\rightarrow$B} & \multicolumn{2}{c|}{I$\rightarrow$C} & \multicolumn{2}{c|}{I$\rightarrow$S} & \multicolumn{2}{c|}{S$\rightarrow$B} & \multicolumn{2}{c|}{S$\rightarrow$C} & \multicolumn{2}{c|}{S$\rightarrow$I} & \multicolumn{2}{c|}{AVG.} \\ \cline{2-15}
			\multicolumn{1}{c|}{} &
			CS (10) & OS (10) & CS (10) & OS (10) & CS (10) & OS (10) & CS (10) & OS (10) & CS (10) & OS (10) & CS (10) & OS (10) & CS (10) & OS (10) \\ \hhline{|=|=|=|=|=|=|=|=|=|=|=|=|=|=|=|}
			LSVM & \textbf{59.1$\pm$2.0} & 42.7$\pm$2.0 & 86.2$\pm$2.6 & 73.3$\pm$3.9 & 50.1$\pm$4.0 & 32.1$\pm$3.2 & 33.1$\pm$1.7 & 16.4$\pm$1.1 & 53.1$\pm$2.6 & 27.9$\pm$2.9 & 52.3$\pm$1.8 & 25.2$\pm$0.5 & 59.3 & 41.5 \\ \hhline{|=|=|=|=|=|=|=|=|=|=|=|=|=|=|=|}
			TCA~\cite{DA_Pan09} & 56.1$\pm$3.8 & 40.9$\pm$2.9 & 83.4$\pm$3.2 & 68.6$\pm$1.8 & 49.3$\pm$2.6 & 34.5$\pm$3.8 & 30.6$\pm$1.3 & 19.4$\pm$2.1 & 47.5$\pm$3.5 & 32.0$\pm$3.9 & 45.2$\pm$1.9 & 31.1$\pm$4.6 & 55.2 & 42.0 \\ \hline
			gfk~\cite{DA_Gong12} & 58.7$\pm$1.9 & 42.6$\pm$2.4 & 86.1$\pm$2.7 & 73.3$\pm$3.6 & 49.5$\pm$3.6 & 32.7$\pm$3.6 & 33.3$\pm$1.4 & 16.9$\pm$1.5 & 53.1$\pm$3.0 & 28.6$\pm$3.8 & 52.5$\pm$2.0 & 26.4$\pm$1.1 & 59.0 & 41.6 \\ \hline 
			SA~\cite{DA_Fernando13} & 58.7$\pm$1.8 & 43.1$\pm$1.6 & 85.9$\pm$2.9 & 72.8$\pm$3.1 & 50.0$\pm$3.6 & 32.2$\pm$3.7 & 34.2$\pm$1.1 & 17.5$\pm$1.6 & 52.5$\pm$3.2 & 29.2$\pm$4.2 & 52.6$\pm$2.4 & 27.1$\pm$1.3 & 59.0 & 41.1 \\ \hline
			CORAL~\cite{CNN-DA_Sun15} & 58.5$\pm$2.7 & 44.6$\pm$2.5 & 85.8$\pm$1.5 & 74.5$\pm$3.4 & 49.5$\pm$4.8 & 35.4$\pm$4.4 & 32.9$\pm$1.6 & 18.7$\pm$1.2 & 52.1$\pm$2.8 & 33.6$\pm$5.3 & 52.9$\pm$1.8 & 31.3$\pm$1.3 & 59.0 & 44.2 \\ \hhline{|=|=|=|=|=|=|=|=|=|=|=|=|=|=|=|}
			ATI & 57.9$\pm$1.9 & \textbf{48.8$\pm$2.3} & 89.3$\pm$2.2 & 77.1$\pm$2.6 & 55.0$\pm$5.0 & 42.2$\pm$4.0 & \textbf{34.9$\pm$2.6} & 22.8$\pm$3.1 & 59.8$\pm$1.3 & 46.9$\pm$2.5 & \textbf{60.8$\pm$3.4} & 32.9$\pm$2.2 & 63.4 & 49.5 \\ \hline
			ATI-$\lambda$ & 58.6$\pm$1.4 & 48.7$\pm$1.8 & \textbf{89.7$\pm$2.3} & \textbf{77.5$\pm$2.2} & \textbf{55.3$\pm$4.3} & \textbf{43.4$\pm$4.8} & 34.1$\pm$2.4 & \textbf{23.2$\pm$3.2} & \textbf{60.2$\pm$2.7} & \textbf{47.3$\pm$2.9} & 60.3$\pm$2.4 & \textbf{33.0$\pm$1.1} & \textbf{63.6} & \textbf{50.2} \\ \hline
		\end{tabular}
	\end{center}	
	\caption{Unsupervised open set domain adaptation on the Testbed dataset (dense setting) with 10 shared classes (OS). In addition, the results for closed set domain adaptation (CS) are reported for comparison.  
	}
	\label{table:testbed}
\end{table*}

In order to measure the performance of our method and the open set protocol across popular datasets with more intra-class variation, we also conduct experiments on the \emph{dense} set-up of the \emph{Testbed for Cross-Dataset Analysis}~\cite{DATA_Tommasi14}.
This protocol provides 40 classes from 4 well known datasets, \emph{Bing (B)}, \emph{Caltech256 (C)}, \emph{ImageNet (I)} and \emph{Sun (S)}. While the samples from the first 3 datasets are mostly centred and without occlusions, \emph{Sun} becomes more challenging due to its collection of object class instances from cluttered scenes. 
As for the Office dataset, we take the first 10 classes as shared classes, the classes 11-25 are used as unknowns in the source domain and the classes 26-40 as unknowns in the target domain. We use the provided DeCAF features (DeCAF7).
Following the unsupervised protocol described in~\cite{DATA_Tommasi15}, we take 50 source samples per class for training and we test on 30 target images per class for all datasets, except \emph{Sun}, where we take 20 samples per class. 

The results reported in Table~\ref{table:testbed} are consistent with the Office dataset. ATI outperforms the baseline and the other methods by +4.1\% for the closed set and by +5.3\% for the open set. 
ATI-$\lambda$ obtains the best accuracies for the open set. 

\subsubsection{Sparse Cross-Dataset Analysis} 

\begin{table}[t]
	\scriptsize
	\begin{center}
		\setlength{\tabcolsep}{.375em}
		\begin{tabular}{|l|c|c|c|c|c|c|c|}
			\cline{2-8}
			\multicolumn{1}{c|}{} &
			C$\rightarrow$O & C$\rightarrow$P & O$\rightarrow$C & O$\rightarrow$P & P$\rightarrow$C & P$\rightarrow$O & AVG. \\ \hline
			\emph{shared classes} & 8 & 7 & 8 & 4 & 7 & 4 & \\ \hline
			\emph{unknown / all (t)} & 0.52 & 0.30 & 0.90 & 0.81 & 0.54 & 0.78 & \\ \hhline{|=|=|=|=|=|=|=|=|}
			LSVM & 46.3 & 36.1 & 60.8 & 29.7 & 78.8 & 70.1 & 53.6 \\ \hline
			TCA~\cite{DA_Pan09} & 45.2 & 33.8 & 58.1 & 31.1 & 63.4 & 61.1 & 48.8 \\ \hline
			gfk~\cite{DA_Gong12} & 46.4 & 36.2 & 61.0 & 29.7 & 79.1 & \textbf{72.6} & 54.2  \\ \hline
			SA~\cite{DA_Fernando13} & 46.4 & 36.8 & 61.1 & 30.2 & 79.8 & 71.1 & 54.2 \\ \hline
			CORAL~\cite{CNN-DA_Sun15} & 48.0 & 35.9 & 60.2 & 29.1 & 78.9 & 68.8 & 53.5 \\ \hhline{|=|=|=|=|=|=|=|=|}
			ATI & \textbf{51.6} & \textbf{52.1} & 63.1 & 38.8 & 80.6 & 70.9 & 59.5 \\ \hline
			ATI-$\lambda$ & 51.5 & 52.0 & \textbf{63.4} & \textbf{39.1} & \textbf{81.1} & 71.1 & \textbf{59.7} \\ \hline
		\end{tabular}
	\end{center}
	\caption{Unsupervised open set domain adaptation on the sparse set-up from~\cite{DATA_Tommasi14}.}
	\label{table:open_uns}
\end{table}

\begin{table}[t]
	\scriptsize
	\begin{center}
		\setlength{\tabcolsep}{.24em}
		\begin{tabular}{|l|c|c|c|c|c|c|c|}
			\cline{2-8}
			\multicolumn{1}{c|}{} &
			C$\rightarrow$O & C$\rightarrow$P & O$\rightarrow$C & O$\rightarrow$P & P$\rightarrow$C & P$\rightarrow$O & AVG. \\ \hline
			LSVM (s) & 46.5$\pm$0.1 & 36.2$\pm$0.1 & 60.8$\pm$0.3 & 29.7$\pm$0.0 & 79.5$\pm$0.3 & 73.5$\pm$0.7 & 54.4 \\ \hline
			LSVM (t) & 53.1$\pm$3.7 & 44.6$\pm$2.1 & 73.7$\pm$1.5 & 40.5$\pm$3.0 & 81.1$\pm$2.5 & 70.5$\pm$4.3 & 60.6 \\ \hline
			LSVM (st) & 56.0$\pm$1.3 & 44.5$\pm$1.2 & 68.9$\pm$1.1 & 40.9$\pm$2.2 & 80.9$\pm$0.6 & 76.7$\pm$0.3 & 61.3 \\ \hhline{|=|=|=|=|=|=|=|=|}			
			ATI & 59.6$\pm$1.2 & 55.2$\pm$1.3 & 75.8$\pm$1.2 & 45.2$\pm$1.4 & 81.6$\pm$0.2 & \textbf{77.1$\pm$0.8} & 65.8 \\ \hline
			ATI-$\lambda$ & 60.3$\pm$1.2 & 56.0$\pm$1.2 & 75.8$\pm$1.1 & \textbf{45.8$\pm$1.2} & 81.8$\pm$0.2 & 76.9$\pm$1.3 & 66.1 \\ \hline
			ATI-$\lambda$-N1 & \textbf{60.7$\pm$1.2} & \textbf{56.3$\pm$1.2} & \textbf{76.7$\pm$1.6} & \textbf{45.8$\pm$1.4} & \textbf{82.0$\pm$0.4} & 76.7$\pm$1.1 & \textbf{66.4} \\ \hline
		\end{tabular}
	\end{center}
	\caption{Semi-supervised open set domain adaptation on the sparse set-up from~\cite{DATA_Tommasi14} with 3 labelled target samples per shared class.}
	\label{table:open_supOffice}
\end{table}

We also introduce an open set evaluation using the \emph{sparse} set-up from~\cite{DATA_Tommasi14} with the datasets \emph{Caltech101 (C)}, \emph{Pascal07 (P)} and \emph{Office (O)}.
These datasets are quite unbalanced and offer distinctive characteristics: \emph{Office} contains centred class instances with barely any background (17 classes, 2300 samples in total, 68-283 samples per class), \emph{Caltech101} allows for more class variety (35 classes, 5545 samples in total, 35-870 samples per class) and \emph{Pascal07} gathers more realistic scenes with partially occluded objects in various image locations (16 classes, 12219 samples in total, 193-4015 samples per class). For each domain shift, we take all samples of the shared classes and consider all other samples as unknowns. Table~\ref{table:open_uns} summarises the amount of shared classes for each shift and the percentage of unknown target samples, which varies from 30\% to 90\%.  


\vspace{1mm}\noindent{\bf Unsupervised domain adaptation.}
For the unsupervised experiment, we conduct a single run for each domain shift using all source and unlabelled target samples. The results are reported in Table~\ref{table:open_uns}. ATI outperforms the baseline and the other methods by +5.3\% for this highly unbalanced open set protocol. ATI-$\lambda$ improves the accuracy of ATI slightly. 

\vspace{1mm}\noindent{\bf Semi-supervised domain adaptation.}
In order to evaluate the semi-supervised setting, we take all source samples and 3 annotated target samples per shared class as it is done in the semi-supervised setting for the Office dataset~\cite{DA_Saenko10}. The average and standard deviation over 5 random splits are reported in Table~\ref{table:open_supOffice}. 
While ATI improves over the baseline trained on the source and target samples together (st) by +4.5\%, ATI-$\lambda$ and the locality constraints with one neighbour boost the performance further. ATI-$\lambda$-$N_1$ improves the accuracy of the baseline by +5.1\%.

\subsubsection{Action recognition}

We extend the applicability of our technique to the field of action recognition in video sequences.
We introduce an open set domain adaptation protocol between the \emph{Kinetics Human Action Video Dataset}~\cite{Data_Kay17} (Kinetics) and the \emph{UCF101 Action Recognition Dataset}~\cite{Data_Soomro12} (UCF101). The Kinects dataset is used as source domain and contains a total of 400 human action classes. The UCF101 dataset serves as target domain including 101 action categories, mainly of sports events. Since the labels of the same action differ between the datasets, \eg, \emph{massaging persons head} (Kinetics) and \emph{head massage} (UCF101), we manually map the class labels between the datasets. Additionally, we also merge all action classes in one datasets if they correspond to a single class in the other dataset, \eg, \emph{dribbling basketball}, \emph{playing basketball}, \emph{shooting basketball} (Kinetics) are merged and associated to \emph{basketball} (UCF101). We finally obtain an open set protocol with 66 shared action classes. 
The list of shared classes, as well as all unrelated categories between both datasets, are provided in the supplemental material. 

For action recognition, we use the features extracted from the 5c layer of the spatial and temporal stream of the I3D model~\cite{3DCNN_Carreira17}, which is pretrained on Kinetics~\cite{Data_Kay17}. We forward the complete video sequences through the spatial and temporal stream of I3D~\cite{3DCNN_Carreira17} and the 5c layer of each stream provides an $7 \times 7 \times 1024$ output for a temporal fragment. We then apply spatial average pooling using a $7\times 7$ kernel and average over time to obtain a $1024$-dimensional feature vector from both the spatial and temporal stream of the I3D model~\cite{3DCNN_Carreira17}. Finally, the feature vectors from the spatial and temporal streams are concatenated to get a single $2048$-dimensional feature vector per video sequence.

\begin{figure}[t]
	\centering
	\subfigure[No adaptation (LSVM): 64.9$\%$]{
		\includegraphics[width=0.474\linewidth]{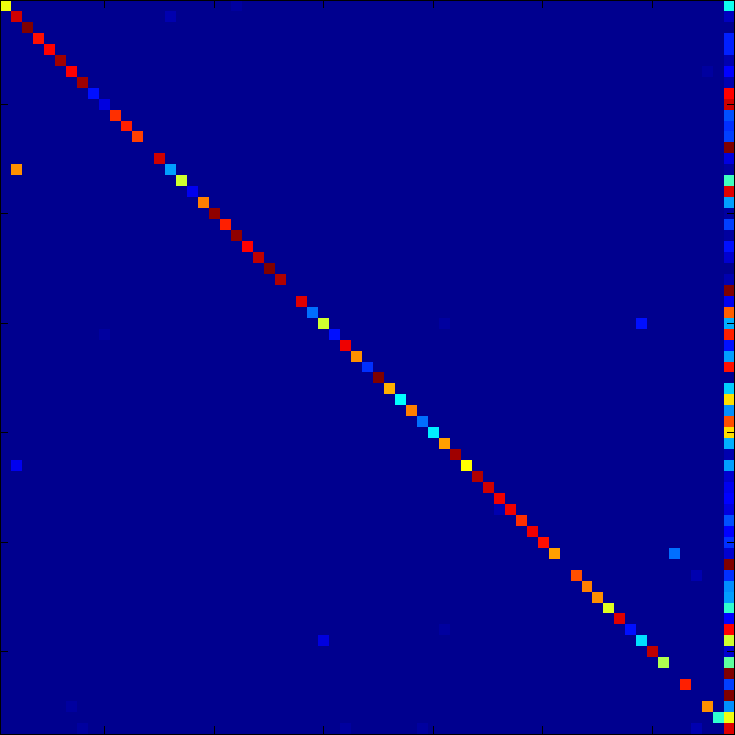}
		\label{fig:ar_lsvm}
	}
	\subfigure[ATI-$\lambda$: 76.9$\%$]{
		\includegraphics[width=0.474\linewidth]{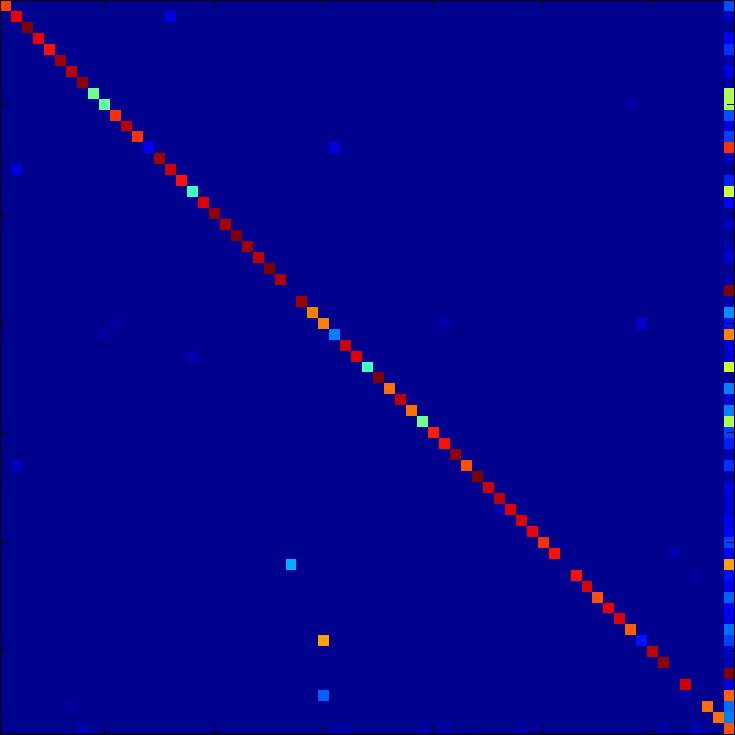}
		\label{fig:ar_da}
	}
	\caption{Confusion matrices without (a) and with adaptation (b) for the 66 shared classes and unknowns (last row and last column) for the open set protocol for \emph{Kinetics}~\cite{Data_Kay17} and \emph{UCF101}~\cite{Data_Soomro12}. Many instances of the shared classes in the target domain are wrongly classified as unknown instances (last column) if domain adaptation is not applied. The figure is best viewed by zooming in.
	}
	\label{fig:ar}
\end{figure}

\vspace{1mm}\noindent{\bf Unsupervised domain adaptation.}
In the unsupervised setting, we evaluate our method by taking all source samples in a single run. Table~\ref{table:ar} shows that the proposed approach outperforms the baseline and other approaches. ATI-$\lambda$ achieves the highest accuracy and improves the accuracy by +12.0$\%$ compared to LSVM. The resulting confusion matrices of LSVM and ATI-$\lambda$ are shown in Fig.~\ref{fig:ar}. LSVM misclassifies many instances of shared classes in the target domain as unknown instances (last column of confusion matrix), which is a well-known problem for open set recognition. Although ATI-$\lambda$ does not resolve this problem completely, it reduces this effect substantially.        


\begin{table}[t]
	\scriptsize
	\begin{center}
		\setlength{\tabcolsep}{.5em}
		\begin{tabular}{|c|c|c|c|c|c|c|}
			\hline
			\multicolumn{7}{|c|}{Kinetics $\rightarrow$ UCF101} \\ \hline
			LSVM & TCA~\cite{DA_Pan09} & gkf~\cite{DA_Gong12} & SA~\cite{DA_Fernando13} & CORAL~\cite{CNN-DA_Sun15} & ATI & ATI-$\lambda$ \\ \hline
			64.9 & 71.2 & 64.9 & 65.1 & 69.4 & 76.6 & \textbf{76.9} \\ \hline
		\end{tabular}
	\end{center}
	\caption{Unsupervised open set domain adaptation for action recognition.}
	\label{table:ar}
\end{table}

\vspace{1mm}\noindent{\bf Semi-supervised domain adaptation.}
We extend the unsupervised protocol to evaluate our method on a semi-supervised setting by labelling 3 target samples per shared class.
We report the average accuracies of 5 random splits in Table~\ref{table:ar_sup}.
Like in the previous semi-supervised experiments, ATI-$\lambda$-N1 obtains the best classification accuracies, outperforming the baseline without adaptation, LSVM (st), by +11.0\%.

\begin{table}[t]
	\scriptsize
	\begin{center}
		\setlength{\tabcolsep}{.5em}
		\begin{tabular}{|c|c|c|c|}
			\hline
			\multicolumn{4}{|c|}{Kinetics $\rightarrow$ UCF101} \\ \hline
			LSVM (st) & ATI & ATI-$\lambda$ & ATI-$\lambda$-N1 \\ \hline
			73.5$\pm$0.5 & 84.1$\pm$0.7 & 84.2$\pm$0.8 & \textbf{84.5$\pm$0.6} \\ \hline
		\end{tabular}
	\end{center}
	\caption{Semi-supervised open set domain adaptation for action recognition.}
	\label{table:ar_sup}
\end{table}

\subsubsection{Synthetic data}

We also introduce another open set protocol with a domain shift between synthetic and real data. In this case, we take 152,397 synthetic images of the VISDA'17 challenge~\cite{DATA_Peng17} as source domain and 5970 instances of real images from the training data of the Pascal3D dataset~\cite{Data_Xiang14} as target domain. Since both datasets contain several types of vehicles, we obtain 6 shared classes, namely, \emph{aeroplane}, \emph{bicycle}, \emph{bus}, \emph{car}, \emph{motorbike} and \emph{train}, within the 12 categories of each dataset.
Following the protocol used in Section~\ref{exp:office}, we extract deep features from the fully connected layer-7 (fc7) from the AlexNet model~\cite{CNN_Krizhevsky12} with 4096 dimensions. In addition, we also extract features from the VGG-16 model~\cite{SimonyanZ14a} to evaluate the impact of using deeper features.

\begin{figure}[t]
	\centering
	\subfigure[No adaptation (LSVM): 53.6$\%$]{
		\includegraphics[width=0.512\linewidth]{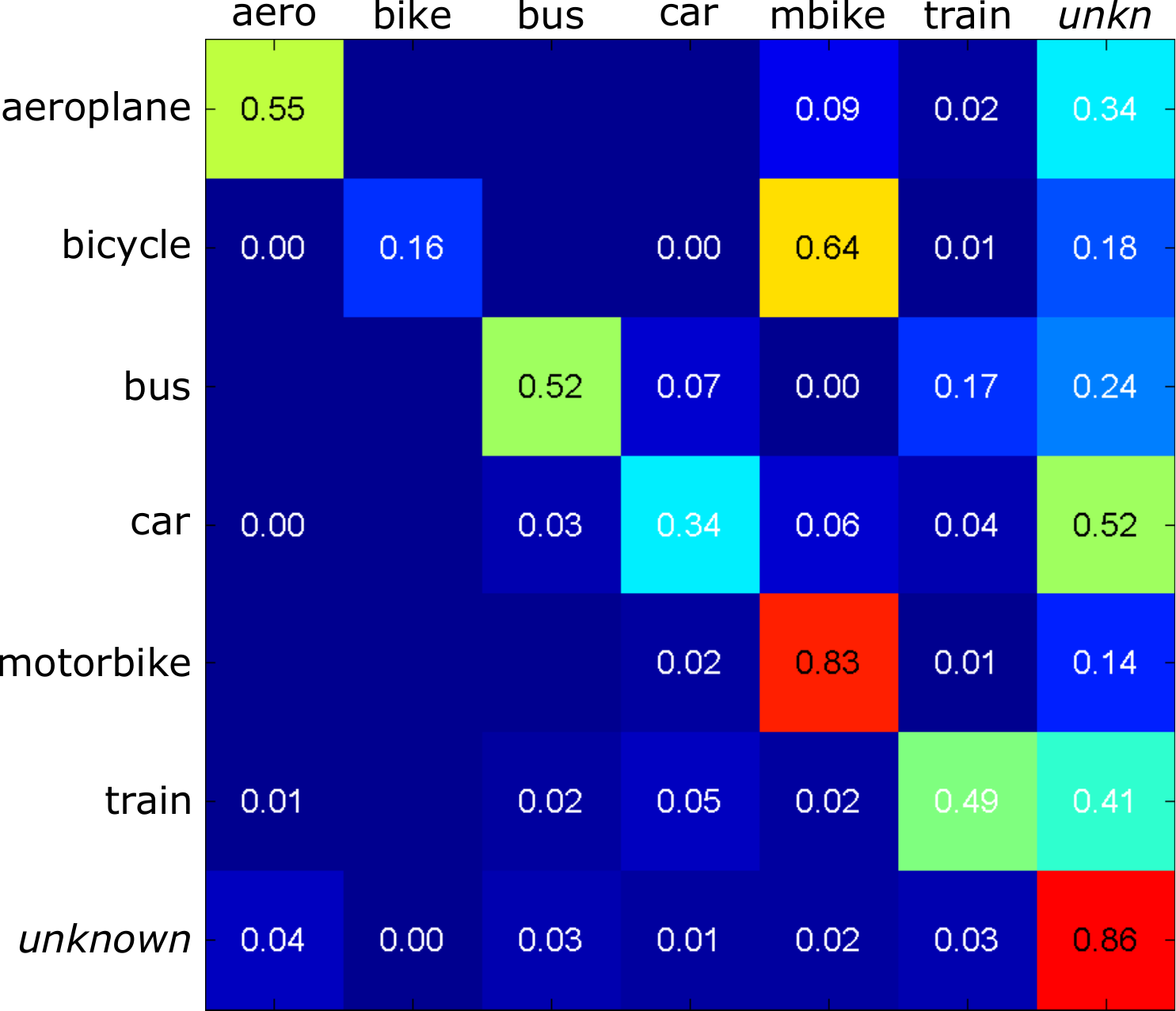}
		\label{fig:syn_lsvm}
	}
	\subfigure[ATI-$\lambda$: 71.9$\%$]{
		\includegraphics[width=0.4225\linewidth]{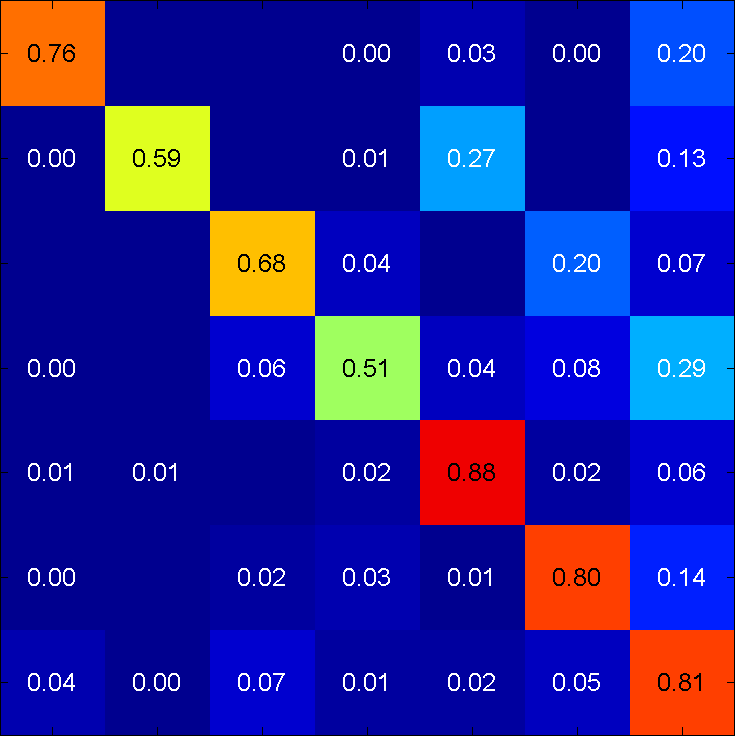}
		\label{fig:syn_da}
	}
	\caption{Confusion matrices without (a) and with adaptation (b) for an open set classification task with 6 shared classes and a domain shift between synthetic~\cite{DATA_Peng17} (source) and real~\cite{Data_Xiang14}  (target) data. The features are extracted from the fc7 layer of the VGG-16 model~\cite{SimonyanZ14a}.
	}
	\label{fig:syn}
\end{figure}

The results of the classification task are shown in Table~\ref{table:syn}. The proposed domain adaptation method achieves the best results for both types of CNN features. 
When we compare the performance of the deep features from AlexNet and VGG-16, the accuracy of the baseline (LSVM) increases by +5.6$\%$ when using the deeper network VGG-16 instead of AlexNet. ATI and ATI-$\lambda$, however, benefit even more from the deeper architecture. For instance, the accuracy of ATI-$\lambda$ increases by +10.5$\%$. This coincides with the observation that deeper networks have a stronger linearisation effect on manifolds of image domains~\cite{BengioMDR13,UpchurchGPPSBW17} than shallow networks. Since the proposed approach learns a linear mapping from the feature space of the source domain to the feature space of the target domain, it benefits from a better linearisation.    
The confusion matrices of the classification task with features extracted from the VGG-16 model are shown in Fig.~\ref{fig:syn}. ATI-$\lambda$ improves the overall accuracy of LSVM by +18.3$\%$ since it resolves confusions between similar classes. For instance, LSVM frequently misclassifies \emph{bicycle} as \emph{motorbike} and \emph{car} as instances of trucks, which are part of the unknown class.

\begin{table}[t]
	\scriptsize
	\begin{center}
		\setlength{\tabcolsep}{.24em}
		\begin{tabular}{|c|c|c|c|c|c|c|c|}
			\cline{2-8}
			\multicolumn{1}{c|}{} & \multicolumn{7}{c|}{VISDA $\rightarrow$ Pascal3D} \\ \cline{2-8}
			\multicolumn{1}{c|}{} & LSVM & TCA~\cite{DA_Pan09} & gkf~\cite{DA_Gong12} & SA~\cite{DA_Fernando13} & CORAL~\cite{CNN-DA_Sun15} & ATI & ATI-$\lambda$ \\ \hline
			AlexNet & 48.0 & 49.7 & 50.1 & 51.2 & 52.0 & 61.1 & \textbf{61.4} \\ \hline
			VGG-16 & 53.6 & 55.0 & 55.2 & 56.5 & 60.0 & \textbf{72.0} & 71.9 \\ \hline
		\end{tabular}
	\end{center}
	\caption{Open set domain adaptation using synthetic images from the VISDA'17 challenge~\cite{DATA_Peng17} as source and real data from the Pascal3D dataset~\cite{Data_Xiang14} as target dataset. There are 6 shared classes between both datasets.}
	\label{table:syn}
\end{table}

\subsection{Closed set domain adaptation}

We also report the accuracies of our method for popular domain adaptation datasets using the standard closed set protocols, where all classes are known in both domains. 

\subsubsection{Office dataset}

For the \emph{Office} dataset~\cite{DA_Saenko10}, we run experiments for the 6 domain shifts of the three provided datasets and use deep features extracted from the fc7 feature map from the AlexNet~\cite{CNN_Krizhevsky12} and VGG-16~~\cite{SimonyanZ14a} models.

\vspace{1mm}\noindent{\bf Unsupervised domain adaptation.}
For unsupervised domain adaptation, we first report the results for the protocol from~\cite{DA_Saenko10}, where we run 5 experiments for each domain shift using randomised samples of the source dataset.
The results are shown in Table~\ref{table:office_uns}, where we compare our method with generic domain adaptation methods, \ie, TCA~\cite{DA_Pan09}, gfk~\cite{DA_Gong12}, SA~\cite{DA_Fernando13} and CORAL~\cite{CNN-DA_Sun15} using AlexNet features.
The results are in accordance with the observations from Section~\ref{exp:office}. While ATI outperforms all generic domain adaptation methods in average and ATI-$\lambda$ performs slightly better than ATI,  
ATI-$\lambda$-$N_1$ decreases the accuracy in the unsupervised setting.
{In addition, we also include the accuracies of using nearest neighbours without domain adaptation, NN, which reports significant lower accuracies than LSVM. LSVM also outperforms NN in other closed set evaluation protocols by a large margin.}

\begin{table}[h]
	\scriptsize
	\begin{center}
		\setlength{\tabcolsep}{.18em}
		\begin{tabular}{|l|c|c|c|c|c|c|c|}
			\cline{2-8}
			\multicolumn{1}{c|}{} &
			A$\rightarrow$D & A$\rightarrow$W & D$\rightarrow$A & D$\rightarrow$W & W$\rightarrow$A & W$\rightarrow$D & AVG. \\ \hline
			NN & 51.3$\pm$1.4 & 45.7$\pm$2.1 & 26.0$\pm$0.9 & 65.5$\pm$1.4 & 28.0$\pm$0.5 & 69.8$\pm$1.8 & 47.7 \\ \hline
			LSVM & 62.3$\pm$3.8 & 55.8$\pm$3.1 & 42.8$\pm$1.6 & 90.1$\pm$0.6 & 41.2$\pm$0.4 & 92.6$\pm$1.5 & 64.1 \\ \hhline{|=|=|=|=|=|=|=|=|}
			TCA~\cite{DA_Pan09} & 60.3$\pm$4.0 & 54.7$\pm$3.0 & 49.4$\pm$1.6 & 90.7$\pm$0.4 & 46.9$\pm$2.3 & 92.0$\pm$0.9 & 65.7 \\ \hline
			gfk~\cite{DA_Gong12} & 61.3$\pm$3.7 & 55.7$\pm$3.0 & 45.6$\pm$1.6 & 90.6$\pm$0.4 & 43.1$\pm$2.3 & 93.4$\pm$0.9 & 65.0 \\ \hline
			SA~\cite{DA_Fernando13} & 60.6$\pm$3.5 & 55.0$\pm$3.1 & 47.3$\pm$1.6 & 90.9$\pm$0.6 & 44.4$\pm$1.4 & 93.3$\pm$0.8 & 65.3 \\ \hline
			CORAL~\cite{CNN-DA_Sun15} & 64.4$\pm$3.9 & 58.9$\pm$3.3 & 52.1$\pm$1.2 & \textbf{92.6$\pm$0.3} & 50.0$\pm$1.0 & \textbf{94.0$\pm$0.6} & 68.7 \\ \hhline{|=|=|=|=|=|=|=|=|}
			ATI & \textbf{67.6$\pm$3.0} & 62.3$\pm$3.1 & 54.8$\pm$1.3 & 90.3$\pm$0.8 & 52.4$\pm$2.1 & 92.6$\pm$1.7 & 70.0 \\ \hline
			ATI-$\lambda$ & 67.3$\pm$2.3 & \textbf{62.6$\pm$2.5} & \textbf{55.2$\pm$2.6} & 90.1$\pm$0.6 & \textbf{53.4$\pm$2.5} & 92.7$\pm$2.5 & \textbf{70.2} \\ \hline
			ATI-$\lambda$-$N_1$ & 64.6$\pm$2.9 & 60.9$\pm$1.3 & 51.9$\pm$1.9 & 90.2$\pm$0.9 & 48.1$\pm$1.6 & 93.7$\pm$2.1 & 68.2 \\ \hline
			
		\end{tabular}
	\end{center}
	\caption{Comparison on the unsupervised Office dataset~\cite{DA_Saenko10} with 31 shared classes and 6 domain shifts using the protocol from~\cite{DA_Saenko10} and features from the AlexNet model (fc7 layer).}
	\label{table:office_uns}
\end{table}

\begin{table}[h]
	\scriptsize
	\begin{center}
		\setlength{\tabcolsep}{.375em}
		\begin{tabular}{|l|c|c|c|c|c|c|c|}
			\cline{2-8}
			\multicolumn{1}{c|}{} &
			A$\rightarrow$D & A$\rightarrow$W & D$\rightarrow$A & D$\rightarrow$W & W$\rightarrow$A & W$\rightarrow$D & AVG. \\ \cline{2-8}
			\multicolumn{1}{c|}{} & \multicolumn{7}{c|}{AlexNet features (fc7)} \\ \hline
			NN & 55.9 & 49.7 & 27.4 & 75.3 & 31.5 & 86.2 & 54.3 \\ \hline
			LSVM & 65.7 & 60.3 & 43.2 & 94.7 & 44.0 & 98.9 & 67.8 \\ \hhline{|=|=|=|=|=|=|=|=|}
			DAN~\cite{CNN-DA_Long15} & 66.8 & 68.5 & 50.0 & 96.0 & 49.8 & 99.0 & 71.7 \\ \hline
			DAH~\cite{CNN-DA_Venkateswara17} & 66.5 & 68.3 & 55.5 & 96.1 & 53.0 & 98.8 & 73.0 \\ \hline
			RTN~\cite{CNN-DA_Long16} & \textbf{71.0} & 73.3 & 50.5 & 96.8 & 51.0 & \textbf{99.6} & 73.7 \\ \hline
			BP~\cite{CNN-DA_Ganin15} & - & 73.0 & - & 96.4 & - & 99.2 & - \\ \hline
			ADDA~\cite{CNN-DA_Tzeng17} & - & \textbf{75.1} & - & \textbf{97.0} & - & \textbf{99.6} & - \\ \hhline{|=|=|=|=|=|=|=|=|}
			ATI & 70.3 & 68.7 & 55.3 & 95.0 & \textbf{56.9} & 98.7 & \textbf{74.2} \\ \hline
			ATI-$\lambda$ & 69.0 & 67.0 & \textbf{56.2} & 95.0 & \textbf{56.9} & 98.7 & 73.8 \\ \hline
			\multicolumn{1}{c|}{} & \multicolumn{7}{c|}{VGG-16 features (fc7)} \\ \hline
			NN & 61.3 & 55.4 & 33.1 & 78.6 & 49.4 & 88.8 & 61.1 \\ \hline
			LSVM & 76.1 & 68.6 & 55.3 & 95.9 & 61.5 & 99.6 & 76.2 \\ \hhline{|=|=|=|=|=|=|=|=|}
			DAN~\cite{CNN-DA_Long15} & 74.4 & 76.0 & 61.5 & 95.9 & 60.3 & 98.6 & 77.8 \\ \hline
			AutoDIAL~\cite{CNN-DA_Carlucci17} & \textbf{82.3} & \textbf{84.2} & 64.6 & \textbf{97.9} & 64.2 & \textbf{99.9} & \textbf{82.2} \\ \hhline{|=|=|=|=|=|=|=|=|}
			ATI & 80.6 & 81.4 & \textbf{67.1} & 96.1 & 66.4 & 99.3 & 81.8 \\ \hline
			ATI-$\lambda$ & 80.8 & 81.3 & 66.9 & 96.1 & \textbf{66.5} & 98.9 & 81.8 \\ \hline
			
		\end{tabular}
	\end{center}
	\caption{Comparison on the unsupervised Office dataset~\cite{DA_Saenko10} with 31 shared classes and 6 domain shifts taking all source samples as in~\cite{DA_Gong13}. 
	}
	\label{table:office_uns_ft_CS}
\end{table}

We also compare our method with current state-of-the-art CNN-based domain adaptation methods \cite{CNN-DA_Long15,CNN-DA_Long16,CNN-DA_Ganin15,CNN-DA_Venkateswara17,CNN-DA_Tzeng17,CNN-DA_Carlucci17}. In this case, we report the accuracies when all source samples are used in a single run as described by~\cite{DA_Gong13}. As shown in Table~\ref{table:office_uns_ft_CS}, our method achieves competitive results even for the standard closed set protocol.    


\begin{table}[h]
	\scriptsize
	\begin{center}
		\setlength{\tabcolsep}{.18em}
		\begin{tabular}{|l|c|c|c|c|c|c|c|}
			\cline{2-8}
			\multicolumn{1}{c|}{} &
			A$\rightarrow$D & A$\rightarrow$W & D$\rightarrow$A & D$\rightarrow$W & W$\rightarrow$A & W$\rightarrow$D & AVG. \\ \cline{2-8}
			\multicolumn{1}{c|}{} & \multicolumn{7}{c|}{AlexNet features (fc7)} \\ \hline
			LSVM (st) & 82.6$\pm$5.5 & 77.0$\pm$2.5 & 63.4$\pm$1.6 & 94.0$\pm$0.8 & 61.8$\pm$1.1 & 96.3$\pm$0.8 & 79.2 \\ \hhline{|=|=|=|=|=|=|=|=|}			
			DDC~\cite{CNN-DA_Tzeng14} & - & 84.1$\pm$0.6 & - & 95.4$\pm$0.4 & - & 96.3$\pm$0.3 & - \\ \hline
			DAN~\cite{CNN-DA_Long15} & - & \textbf{85.7$\pm$0.3} & - & \textbf{97.2$\pm$0.2} & - & 96.4$\pm$0.2 & - \\ \hline 
			MMC~\cite{CNN-DA_Tzeng15} & 86.1$\pm$1.2 & 82.7$\pm$0.8 & \textbf{66.2$\pm$0.3} & 95.7$\pm$0.5 & 65.0$\pm$0.5 & \textbf{97.6$\pm$0.2} & \textbf{82.2} \\
			\hhline{|=|=|=|=|=|=|=|=|}
			ATI (labels t) & 85.0$\pm$2.1 & 78.3$\pm$2.3 & 63.6$\pm$1.5 & 94.0$\pm$0.8 & 62.3$\pm$0.9 & 96.4$\pm$0.8 & 79.9 \\ \hline
			ATI & 85.5$\pm$2.9 & 82.4$\pm$1.1 & 65.1$\pm$1.3 & 93.4$\pm$0.9 & 65.6$\pm$1.5 & 95.7$\pm$1.1 & 81.3 \\ \hline
			ATI-$\lambda$ & 85.6$\pm$2.6 & 82.6$\pm$0.5 & 65.3$\pm$1.3 & 93.3$\pm$1.0 & 65.7$\pm$1.7 & 95.7$\pm$1.1 & 81.4 \\ \hline
			ATI-$\lambda$-$N_1$ & \textbf{88.1$\pm$1.7} & 83.1$\pm$2.3 & 66.0$\pm$1.4 & 93.9$\pm$1.2 & \textbf{65.9$\pm$1.5} & 96.2$\pm$0.8 & \textbf{82.2} \\ \hline
			ATI-$\lambda$-$N_2$ & 87.0$\pm$3.5 & 84.6$\pm$3.5 & 65.3$\pm$1.0 & 93.6$\pm$1.2 & \textbf{65.9$\pm$1.8} & 95.8$\pm$1.3 & 82.0 \\ \hline
			\multicolumn{1}{c|}{} & \multicolumn{7}{c|}{VGG-16 features (fc7)} \\ \hline
			LSVM (st) & 86.1$\pm$1.5 & 83.4$\pm$1.2 & 67.9$\pm$1.0 & 96.1$\pm$0.7 & 67.1$\pm$0.6 & \textbf{96.6$\pm$1.0} & 82.9 \\ \hhline{|=|=|=|=|=|=|=|=|}
			SO~\cite{CNN-DA_Koniusz17} & 84.5$\pm$1.7 & 86.3$\pm$0.8 & 65.7$\pm$1.7 & 97.5$\pm$0.7 & 66.5$\pm$1.0 & 95.5$\pm$0.6 & 82.7 \\ \hline
			CCSA~\cite{CNN-DA_Motiian17} & 88.2$\pm$1.0 & \textbf{89.0$\pm$1.2} & \textbf{72.1$\pm$1.0} & \textbf{97.6$\pm$0.4} & \textbf{71.8$\pm$0.5} & 96.4$\pm$0.8 & \textbf{85.8} \\ \hhline{|=|=|=|=|=|=|=|=|}
			ATI-$\lambda$-$N_1$ & \textbf{90.3$\pm$1.9} & 88.0$\pm$1.4 & 70.8$\pm$0.9 & 95.1$\pm$0.7 & 70.3$\pm$2.0 & 96.3$\pm$0.9 & 85.1 \\ \hline
		\end{tabular}
	\end{center}
	\caption{Comparison on the semi-supervised Office dataset~\cite{DA_Saenko10} with 31 shared classes and 6 domain shifts, following the protocol from~\cite{DA_Saenko10}.  
	}
	\label{table:office_sup_cs}
\end{table}

\vspace{1mm}\noindent{\bf Semi-supervised domain adaptation.}
We also evaluate our approach for semi-supervised domain adaptation on the \emph{Office} dataset. We follow the protocol from~\cite{DA_Saenko10} and report the accuracies and standard deviations over 5 runs with random samples.
In the first experiment with AlexNet features, we also include ATI-$\lambda$-$N_2$ with locality constraints using 2 nearest neighbours and compare our approach with state-of-the-art CNN-based methods~\cite{CNN-DA_Tzeng14,CNN-DA_Long15,CNN-DA_Tzeng15}. As in Section~\ref{exp:office}, we train the SVMs on the transformed source samples and labelled target samples (st). 
The results are reported in Table~\ref{table:office_sup_cs}.

Our method achieves the same average accuracy as MMC~\cite{CNN-DA_Tzeng15} and performs slightly worse than \cite{CNN-DA_Motiian17} for the VGG-16 features. In addition, we report the accuracy for AlexNet features when the mapping $W$ \eqref{eq:objMat} is estimated using only the labelled target samples without solving the individual assignments \eqref{eq:assignments}. This variant is denoted by ATI (labels t) and performs worse than ATI. 


\subsubsection{Office+Caltech dataset}

\begin{table}[h]
	\scriptsize
	\setlength{\tabcolsep}{.59em}
	\begin{tabular}{|c|c|c|c|c|c|c|}
		\cline{2-7}
		\multicolumn{1}{c|}{} &
		A$\rightarrow$C & A$\rightarrow$D & A$\rightarrow$W & C$\rightarrow$A & C$\rightarrow$D & C$\rightarrow$W \\ \cline{2-7}
		\multicolumn{1}{c|}{} & \multicolumn{6}{c|}{AlexNet features (fc7)} \\ \hline
		NN & 78.4 & 78.1 & 71.7 & 90.7 & 84.4 & 80.8 \\ \hline
		LSVM & 83.3 & 84.1 & 77.5 & 91.8 & 89.1 & 82.3 \\ \hhline{|=|=|=|=|=|=|=|}
		CORAL~\cite{CNN-DA_Sun15} & 83.2 & 86.5 & 79.6 & 91.4 & 86.6 & 82.1 \\ \hline
		BP~\cite{CNN-DA_Ganin15} & 84.6 & 92.3 & 90.2 & 91.9 & 92.8 & 93.2 \\ \hline
		DDC~\cite{CNN-DA_Tzeng14} & 83.5 & 88.4 & 83.1 & 91.9 & 88.8 & 85.4 \\ \hline
		DAN~\cite{CNN-DA_Long15} & 84.1 & 91.1 & 91.8 & 92.0 & 89.3 & 90.6 \\ \hline
		RTN\cite{CNN-DA_Long16} & \textbf{88.1} & \textbf{95.5} & \textbf{95.2} & 93.7 & \textbf{94.2} & \textbf{96.9} \\ \hhline{|=|=|=|=|=|=|=|}
		ATI & 86.5 & 92.8 & 88.7 & \textbf{93.8} & 89.6 & 93.6 \\ \hline
		ATI-$\lambda$ & 87.1 & 90.6 & 90.7 & 93.4 & 85.4 & 93.4 \\ \hline
		\multicolumn{1}{c|}{} & \multicolumn{6}{c|}{VGG-16 features (fc7)} \\ \hline
		NN & 86.7 & 84.4 & 83.4 & 91.4 & 88.2 & 88.0 \\ \hline
		LSVM & 87.8 & 88.7 & 87.2 & 93.3 & 91.8 & 91.4 \\ \hhline{|=|=|=|=|=|=|=|}
		ATI & 91.0 & 92.4 & 95.9 & 94.7 & 93.1 & 97.4\\ \hline
		ATI-$\lambda$ & 90.4 & 92.4 & 91.4 & 94.5 & 93.9 & 96.0 \\ \hline
	\end{tabular}
	\setlength{\tabcolsep}{.56em}
	\begin{tabular}{|c|c|c|c|c|c|c|c|}
		\cline{2-8}
		\multicolumn{1}{c|}{} & D$\rightarrow$A & D$\rightarrow$C & D$\rightarrow$W & W$\rightarrow$A & W$\rightarrow$C & W$\rightarrow$D & AVG \\ \cline{2-8}
		\multicolumn{1}{c|}{} & \multicolumn{7}{c|}{AlexNet features (fc7)} \\ \hline
		NN & 64.2 & 58.6 & 89.0 & 63.2 & 58.8 & 95.4 & 76.1 \\ \hline
		LSVM & 79.4 & 70.2 & 97.9 & 80.0 & 72.7 & \textbf{100.0} & 84.0 \\ \hhline{|=|=|=|=|=|=|=|=|}
		CORAL~\cite{CNN-DA_Sun15} & 87.3 & 77.5 & \textbf{99.3} & 85.2 & 76.1 & \textbf{100.0} & 86.2 \\ \hline
		BP~\cite{CNN-DA_Ganin15} & 84.0 & 74.9 & 97.8 & 86.9 & 77.3 & \textbf{100.0} & 88.2 \\ \hline
		DDC~\cite{CNN-DA_Tzeng14} & 89.0 & 79.2 & 98.1 & 84.9 & 73.4 & \textbf{100.0} & 87.1 \\ \hline
		DAN~\cite{CNN-DA_Long15} & 90.0 & 80.3 & 98.5 & 92.1 & 81.2 & \textbf{100.0} & 90.1 \\ \hline
		RTN\cite{CNN-DA_Long16} & \textbf{93.8} & 84.6 & 99.2 & \textbf{95.5} & \textbf{86.6} & \textbf{100.0} & \textbf{93.4} \\ \hhline{|=|=|=|=|=|=|=|=|}
		ATI & 93.4 & \textbf{85.9} & 98.9 & 93.6 & 86.3 & \textbf{100.0} & 91.9 \\ \hline
		ATI-$\lambda$ & 93.6 & 85.8 & \textbf{99.3} & 93.6 & 86.1 & \textbf{100.0} & 91.8 \\ \hline
		\multicolumn{1}{c|}{} & \multicolumn{7}{c|}{VGG-16 features (fc7)} \\ \hline
		NN & 78.9 & 75.0 & 95.2 & 80.9 & 78.5 & 100.0 & 85.6 \\ \hline
		LSVM & 82.5 & 77.9 & 98.4 & 87.8 & 84.9 & 100.0 & 89.3 \\ \hhline{|=|=|=|=|=|=|=|=|}
		ATI & 93.7 & 89.8 & 98.1 & 95.1 & 90.3 & 99.5 & 94.3 \\ \hline
		ATI-$\lambda$ & 94.6 & 89.4 & 98.4 & 95.3 & 89.4 & 99.6 & 93.8 \\ \hline
	\end{tabular}
	\vspace{1mm}	
	\caption{Classification accuracies on the unsupervised Office+Caltech dataset~\cite{DA_Gong12} with 10 shared classes and 12 domain shifts using deep features. We take all source samples on a single run~\cite{DA_Gong13}. 
	}
	\label{table:saenko_uns_ft}
\end{table}

We also evaluate our approach on the extended version of the Office evaluation set~\cite{DA_Gong12}, which includes the additional \emph{Caltech (C)} dataset. This results in 12 domain shifts, but reduces the amount of shared classes to only 10.
As shown in Table~\ref{table:saenko_uns_ft}, our method obtains very competitive results with AlexNet features, outperforming in overall the generic domain adaptation method~\cite{CNN-DA_Sun15} and 3 out of 4 CNN-based methods. If features from a deeper network such as VGG-16 are used, our method obtains the best overall results.

\subsubsection{Dense Testbed for Cross-Dataset Analysis}

\begin{table}[h]
	\scriptsize
	\setlength{\tabcolsep}{.20em}
	\begin{tabular}{|c|c|c|c|c|c|c|}
		\cline{2-7}
		\multicolumn{1}{c|}{} &
		B$\rightarrow$C & B$\rightarrow$I & B$\rightarrow$S & C$\rightarrow$B & C$\rightarrow$I & C$\rightarrow$S \\ \hline
		LSVM & 63.8$\pm$2.2 & 57.4$\pm$0.7 & 20.2$\pm$1.0 & 38.3$\pm$0.8 & 62.9$\pm$0.9 & 21.7$\pm$1.6 \\ \hhline{|=|=|=|=|=|=|=|}
		TCA~\cite{DA_Pan09} & 53.8$\pm$1.3 & 49.1$\pm$1.1 & 17.1$\pm$1.1 & 35.6$\pm$1.8 & 59.2$\pm$0.8 & 18.9$\pm$1.2 \\ \hline
		gfk~\cite{DA_Gong12} & 63.4$\pm$1.8 & 57.2$\pm$1.1 & 20.6$\pm$1.3 & 38.3$\pm$0.9 & 62.9$\pm$1.2 & 21.7$\pm$1.4 \\ \hline
		SA~\cite{DA_Fernando13} & 63.0$\pm$1.9 & 57.1$\pm$1.4 & 20.2$\pm$1.4 & 38.3$\pm$0.9 & 62.8$\pm$1.0 & 21.5$\pm$1.2 \\ \hline
		CORAL~\cite{CNN-DA_Sun15} & 63.9$\pm$2.1 & 57.8$\pm$0.8 & 20.4$\pm$2.0 & 38.3$\pm$0.8 & 63.4$\pm$0.9 & 22.5$\pm$1.2 \\ \hhline{|=|=|=|=|=|=|=|}
		ATI & 69.1$\pm$1.3 & 62.4$\pm$1.9 & 23.4$\pm$1.1 & \textbf{39.0$\pm$1.4} & \textbf{66.9$\pm$1.2} & 25.2$\pm$0.9 \\ \hline
		ATI-$\lambda$ & \textbf{69.4$\pm$1.4} & \textbf{62.9$\pm$1.3} & \textbf{23.6$\pm$1.0} & \textbf{39.0$\pm$1.4} & \textbf{66.9$\pm$1.1} & \textbf{25.3$\pm$0.9} \\ \hline
	\end{tabular}
	\begin{tabular}{|c|c|c|c|c|c|c|c|}
		\cline{2-8}
		\multicolumn{1}{c|}{} & I$\rightarrow$B & I$\rightarrow$C & I$\rightarrow$S & S$\rightarrow$B & S$\rightarrow$C & S$\rightarrow$I & AVG \\ \hline
		LSVM & 39.3$\pm$1.4 & 70.8$\pm$1.5 & 24.6$\pm$1.8 & 16.6$\pm$1.0 & 26.1$\pm$2.0 & 26.3$\pm$0.7 & 39.0 \\ \hhline{|=|=|=|=|=|=|=|=|}
		TCA~\cite{DA_Pan09} & 36.4$\pm$1.2 & 66.3$\pm$2.3 & 22.2$\pm$1.4 & 13.8$\pm$1.4 & 23.2$\pm$1.5 & 23.2$\pm$1.5 & 34.9 \\ \hline
		gfk~\cite{DA_Gong12} & 38.8$\pm$1.3 & 70.9$\pm$1.1 & 24.4$\pm$1.4 & 16.3$\pm$0.9 & 26.7$\pm$1.8 & 26.1$\pm$1.0 & 38.9 \\ \hline
		SA~\cite{DA_Fernando13} & 39.0$\pm$1.3 & 71.1$\pm$1.3 & 24.2$\pm$1.4 & 16.0$\pm$0.9 & 26.8$\pm$1.9 & 26.4$\pm$1.1 & 38.9 \\ \hline
		CORAL~\cite{CNN-DA_Sun15} & 39.0$\pm$1.2 & 71.2$\pm$1.3 & 24.9$\pm$1.6 & 16.8$\pm$1.0 & 27.4$\pm$2.2 & 27.7$\pm$0.5 & 39.4 \\ \hhline{|=|=|=|=|=|=|=|=|}
		ATI & 39.7$\pm$1.8 & 74.4$\pm$1.6 & \textbf{25.9$\pm$2.1} & 18.3$\pm$1.1 & 37.1$\pm$3.2 & \textbf{35.0$\pm$1.0} & 42.8 \\ \hline
		ATI-$\lambda$ & \textbf{39.8$\pm$1.8} & \textbf{74.8$\pm$1.5} & 25.8$\pm$2.0 & \textbf{18.7$\pm$0.7} & \textbf{37.4$\pm$2.9} & 34.8$\pm$0.8 & \textbf{43.2} \\ \hline
	\end{tabular}	
	\vspace{2.5mm}	
	\caption{\emph{Testbed} dataset~\cite{DATA_Tommasi14} with 40 common classes and 12 domain shifts.}
	\label{table:testbed_dense}
\end{table}

We also present an evaluation on the Dense dataset of the Testbed for Cross-Dataset Analysis~\cite{DATA_Tommasi15} using the provided DeCAF features. This protocol comprises 12 domain shifts between the 4 datasets \emph{Bing (B)}, \emph{Caltech (C)}, \emph{ImageNet (I)} and \emph{Sun (S)}, which share 40 classes.
Following the protocol described in~\cite{DATA_Tommasi15}, we take 50 source samples per class for training and we test on 30 target images per class for all datasets, except \emph{Sun}, where we take 20 samples per class.
The results reported in Table~\ref{table:testbed_dense} show that ATI-$\lambda$ outperforms other generic domain adaptation methods. 

\subsubsection{Sentiment Analysis}

\begin{table}[h]
	\scriptsize
	\begin{center}
		\setlength{\tabcolsep}{.375em}
		\begin{tabular}{|l|c|c|c|c|c|}
			\cline{2-6}
			\multicolumn{1}{c|}{} &
			B$\rightarrow$E & D$\rightarrow$B & E$\rightarrow$K & K$\rightarrow$D & AVG. \\ \hline
			LSVM & 75.5$\pm$1.6 & 78.2$\pm$2.5 & 83.1$\pm$1.8 & 73.3$\pm$1.8 & 77.5 \\ \hhline{|=|=|=|=|=|=|}
			TCA~\cite{DA_Pan09} & 76.6$\pm$2.2 & 78.5$\pm$1.6 & \textbf{83.8$\pm$1.5} & 75.0$\pm$1.4 & 78.5 \\ \hline
			gfk~\cite{DA_Gong12} & 77.0$\pm$2.0 & \textbf{79.2$\pm$1.8} & 83.7$\pm$1.7 & 73.7$\pm$1.9 & 78.4 \\ \hline
			SA~\cite{DA_Fernando13} & 75.9$\pm$1.9 & 78.4$\pm$2.1 & 83.0$\pm$1.7 & 72.1$\pm$1.9 & 77.4 \\ \hline
			CORAL~\cite{CNN-DA_Sun15} & 76.2$\pm$1.7 & 78.4$\pm$2.0 & 83.1$\pm$2.0 & 74.2$\pm$3.0 & 78.0 \\ \hhline{|=|=|=|=|=|=|}
			ATI & \textbf{79.9$\pm$2.0} & \textbf{79.2$\pm$1.9} & 83.7$\pm$2.1 & \textbf{75.6$\pm$1.9} & \textbf{79.6} \\ \hline
			ATI-$\lambda$ & 79.6$\pm$1.4 & 79.0$\pm$1.8 & 83.6$\pm$2.1 & 74.4$\pm$1.7 & 79.2 \\ \hline
			
		\end{tabular}
	\end{center}
	\caption{Accuracies of 4 domain shifts on the Sentiment dataset~\cite{DA_Blitzer07} using the bag-of-words features and the protocol from~\cite{DA_Gong13ICML}.}
	\label{table:sentiment}
\end{table}

To show the behaviour of our method with a different type of feature descriptor, we also present an evaluation on the \emph{Sentiment analysis} dataset~\cite{DA_Blitzer07}. This dataset gathers reviews from Amazon for four products: \emph{books (B)}, \emph{DVDs (D)}, \emph{electronics (E)} and \emph{kitchen appliances (K)}. Each domain contains 1000 reviews labelled as \emph{positive} and another set of 1000 reviews as \emph{negative}.
We use the data provided by~\cite{DA_Gong13ICML}, which extracts bag-of-words features from the 400 words with the largest mutual information across domains.
We report the mean accuracy over 20 splits, where for each run 1600 samples are randomly selected for training and the other 400 for testing. The results in Table~\ref{table:sentiment} show that our approach not only works very well for image and video data, but it can also be applied to other types of data. This demonstrates the versatility of the proposed approach.


\section{Conclusions}

We have introduced the concept of open set domain adaptation in the context of image classification and action recognition. In contrast to closed set domain adaptation, we do not assume that all instances in the source and target domain belong to the same set of classes, but allow that each domain contains instances of classes that are not present in the other domain. We furthermore proposed an approach for unsupervised and semi-supervised domain adaptation that achieves state-of-the-art results for open sets and competitive results for closed sets. In particular, the flexibility of the approach, which can be used for images, videos and other types of data, makes the approach a versatile tool for real-world applications.

\ifCLASSOPTIONcompsoc
  \section*{Acknowledgments}
\else
  \section*{Acknowledgment}
\fi

The work has been supported by the ERC Starting Grant ARCA (677650) and the DFG projects GA 1927/2-2 (DFG Research Unit FOR 1505 Mapping on Demand) and GA 1927/4-1 (DFG Research Unit FOR 2535 Anticipating Human Behavior).

\ifCLASSOPTIONcaptionsoff
  \newpage
\fi



\bibliographystyle{IEEEtran}
\bibliography{IEEEabrv,egbib}
%


%

\begin{IEEEbiography}[{\includegraphics[width=1in,height=1.25in,clip,keepaspectratio]{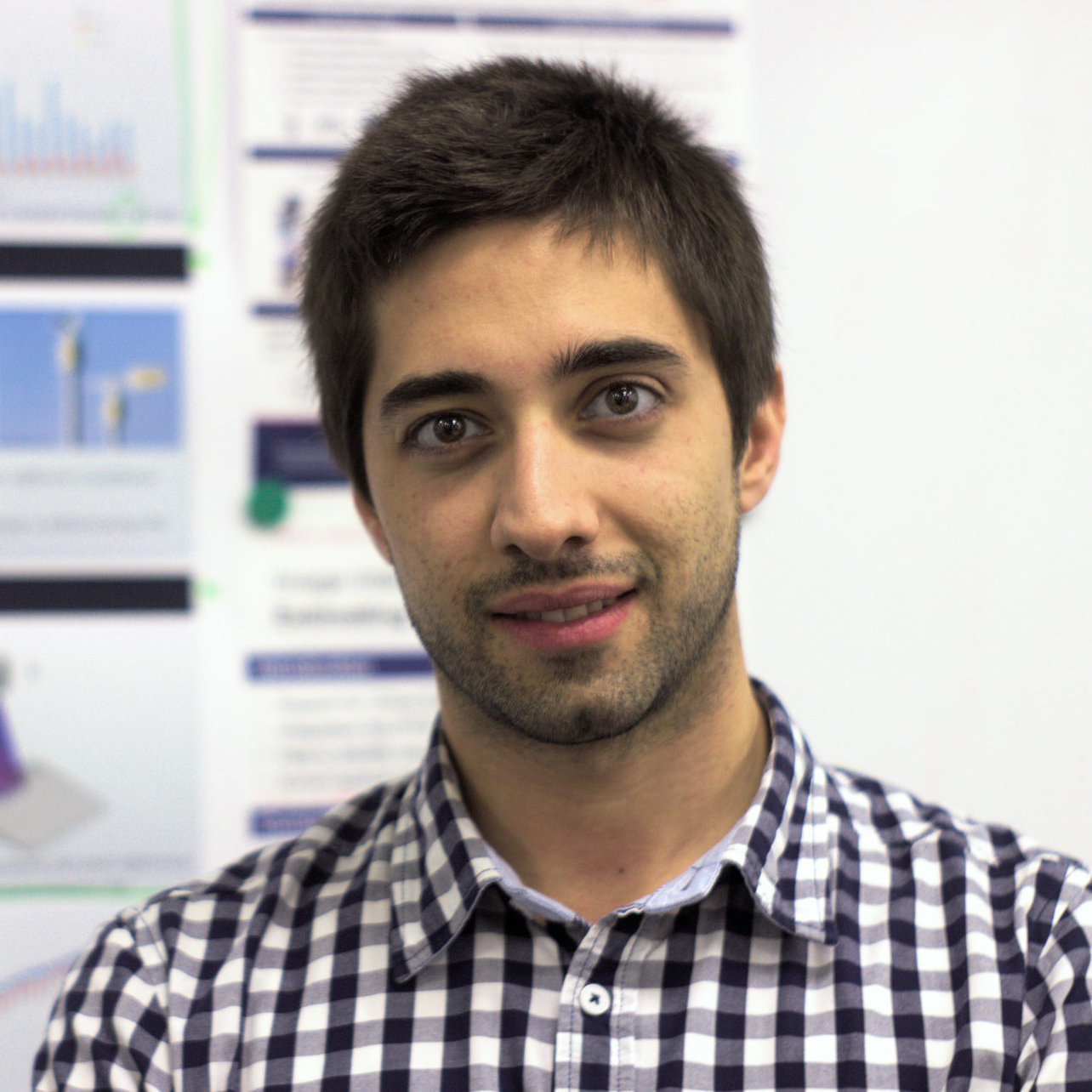}}]{Pau Panareda Busto}
received his B.Sc. and Masters degrees in computer engineering from the Technical University of Catalonia (2010) and his Masters degree in media informatics from the RWTH Aachen Universtiy (2013). Since September 2013, he has been a Ph.D. candidate at the University of Bonn in collaboration with Airbus Group.
His research interests include computer vision, computer graphics and machine learning.
\end{IEEEbiography}

\begin{IEEEbiography}
[{\includegraphics[width=1in,height=1.25in,clip,keepaspectratio]{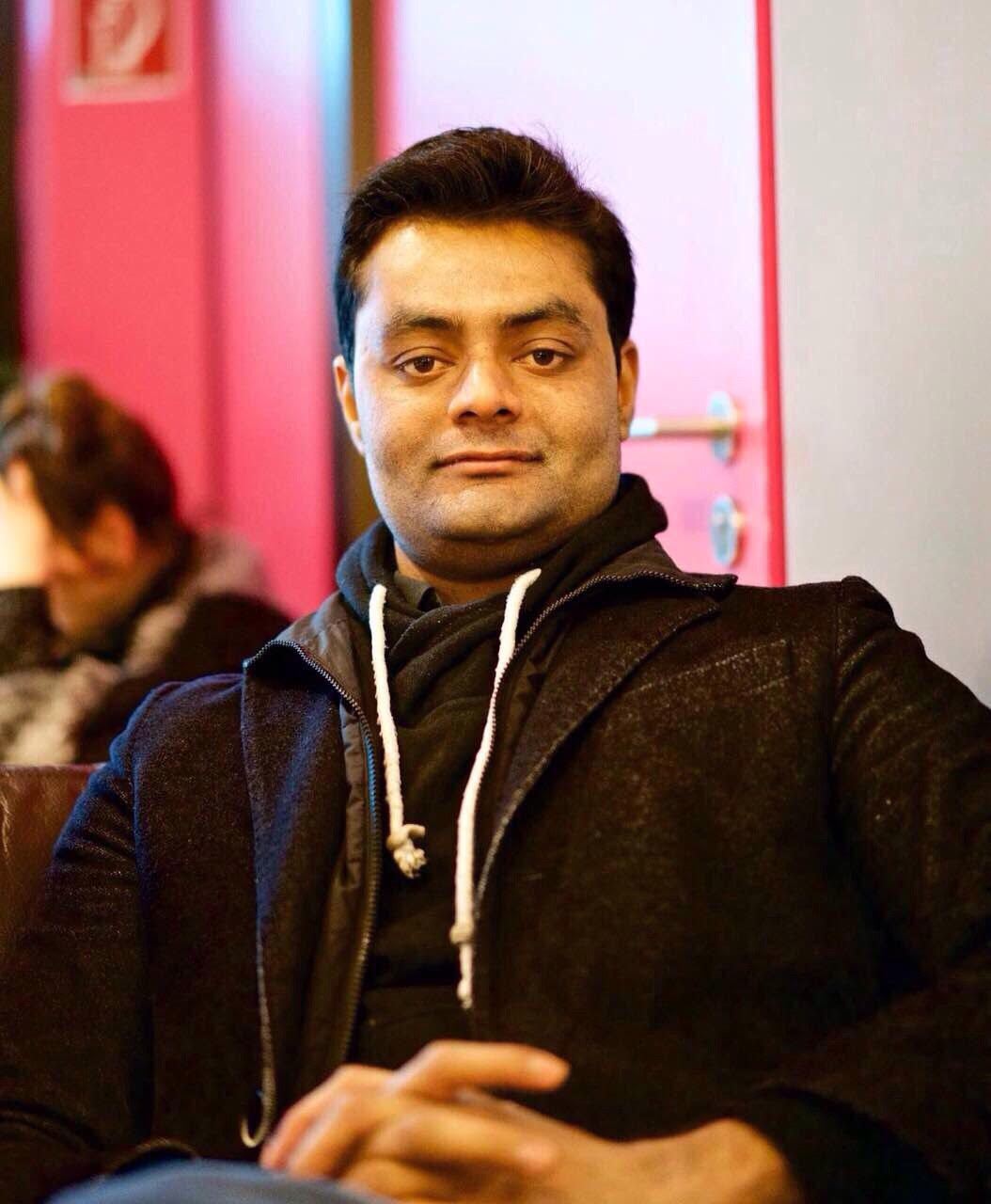}}]{Ahsan Iqbal} obtained his master degree in Computer Science from the University of Bonn in 2017. He did an internship in Amazon Berlin from December 2016 to March 2017. Since April 2017, he is a Ph.D. student at the University of Bonn. His research interests are action recognition and action detection.
\end{IEEEbiography}


\begin{IEEEbiography}
[{\includegraphics[width=1in,height=1.25in,clip,keepaspectratio]{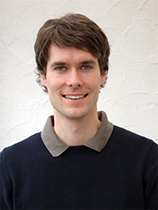}}]{Juergen Gall} obtained his B.Sc. and his Masters degree in
mathematics from the University of Wales Swansea (2004)
and from the University of Mannheim (2005). In 2009,
he obtained a Ph.D. in computer science from the Saarland
University and the Max Planck Institut f{\"u}r Informatik. He
was a postdoctoral researcher at the Computer Vision Laboratory, ETH Zurich,
from 2009 until 2012 and senior research scientist at the Max Planck Institute for Intelligent
Systems in T{\"u}bingen from 2012 until 2013. Since 2013,
he is professor at the University of Bonn and head of the
Computer Vision Group.  
\end{IEEEbiography}




\end{document}